\documentclass[10pt,twocolumn,letterpaper]{article}

\usepackage[pagenumbers]{cvpr} 

\definecolor{cvprblue}{rgb}{0.21,0.49,0.74}
\usepackage[pagebackref,breaklinks,colorlinks,allcolors=cvprblue]{hyperref}

\usepackage{url}
\usepackage{multirow} 
\usepackage{graphicx}
\usepackage[export]{adjustbox}
\usepackage{float}
\usepackage{epsfig}
\usepackage{fontawesome5}
\usepackage{hyperref}
\usepackage{graphicx}
\usepackage{amsmath}
\usepackage{amssymb} 
\usepackage{caption}
\usepackage{xparse} 
\usepackage{wrapfig} 
\usepackage{tabularx}
\usepackage{subcaption}
\usepackage{fancyhdr}
\usepackage{color}
\usepackage{wrapfig} 
\usepackage{framed}
\usepackage{soul}
\usepackage{xspace}
\usepackage{booktabs} 
\usepackage{makecell}
\usepackage{xcolor}
\usepackage{vcell}
\usepackage{colortbl}
\usepackage[nodisplayskipstretch]{setspace}
\usepackage{seqsplit} 
\usepackage{array}
\usepackage{CJKutf8} 
\usepackage[normalem]{ulem}
\usepackage{enumitem}
\usepackage{etoc}
\usepackage{tcolorbox}
\usepackage[misc,geometry]{ifsym}
\usepackage{titletoc}

\definecolor{backred}{RGB}{255, 190, 190}
\definecolor{backblue}{RGB}{210, 230, 250}
\definecolor{myblue}{RGB}{6, 174, 226}

\definecolor{darkgreen}{rgb}{0.0,0.5,0.0}

\definecolor{shadecolor}{RGB}{237,237,237}

\newcommand{\worldwideweb}{\raisebox{-1.5pt}{\includegraphics[height=1.05em]{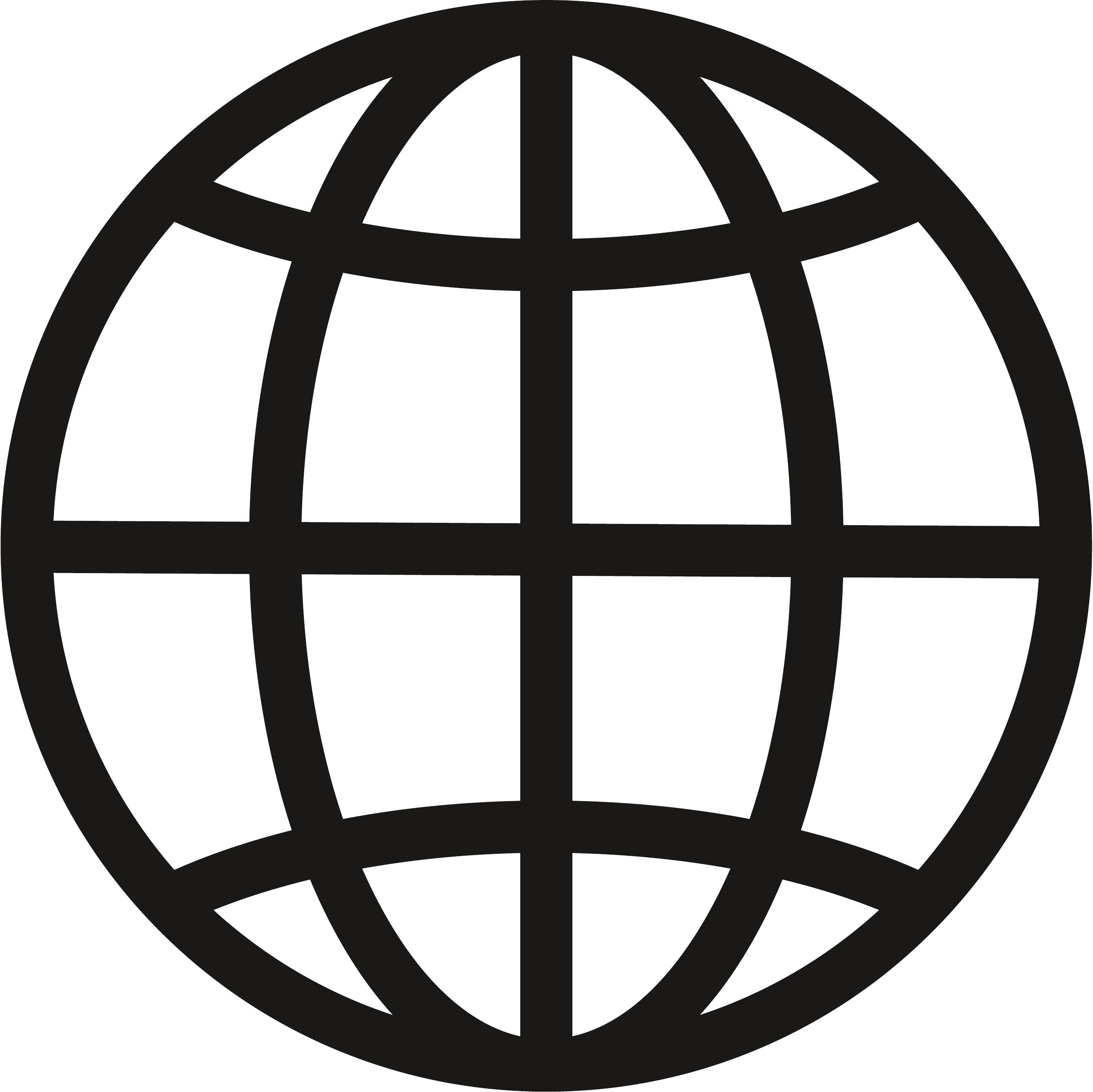}}\xspace}
\newcommand{\github}{\raisebox{-1.5pt}{\includegraphics[height=1.05em]{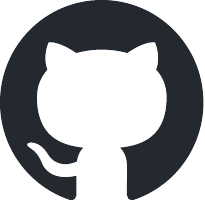}}\xspace}
\newcommand{\huggingface}{\raisebox{-1.5pt}{\includegraphics[height=1.05em]{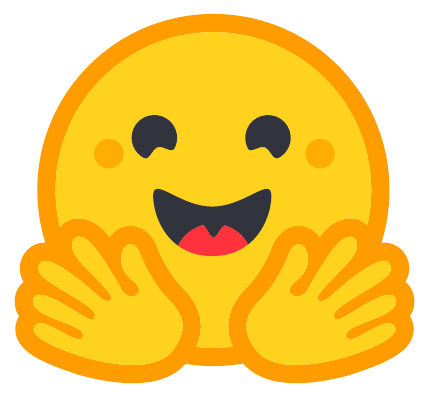}}\xspace}

\definecolor{uclablue}{rgb}{0.15, 0.45, 0.68}
\definecolor{Gray}{gray}{0.93}
\definecolor{uclagold}{rgb}{1.0, 0.7, 0.0}
\definecolor{airforceblue}{rgb}{0.36, 0.54, 0.66}
\definecolor{rosegold}{rgb}{0.72, 0.43, 0.47}
\definecolor{pastelbrown}{rgb}{0.51, 0.41, 0.33}
\definecolor{isabelline}{rgb}{0.96, 0.94, 0.93}
\definecolor{macaroniandcheese}{rgb}{0.98, 0.89, 0.83}
\definecolor{wildblueyonder}{rgb}{0.85, 0.89, 0.95}
\definecolor{mediumtaupe}{rgb}{0.4, 0.3, 0.28}
\definecolor{bluegray}{rgb}{0.4, 0.6, 0.8}
\definecolor{celestialblue}{rgb}{0.29, 0.59, 0.82}
\definecolor{darkorange}{rgb}{1.0, 0.55, 0.0}
\definecolor{cadmiumred}{rgb}{0.89, 0.0, 0.13}
\definecolor{magnolia}{rgb}{0.97, 0.96, 1.0}
\definecolor{pastelblue}{rgb}{0.68, 0.78, 0.81}
\definecolor{persiangreen}{rgb}{0.0, 0.65, 0.58}
\definecolor{steelblue}{rgb}{0.27, 0.51, 0.71}
\definecolor{bluebell}{rgb}{0.64, 0.64, 0.82}
\definecolor{dimgray}{rgb}{0.41, 0.41, 0.41}
\definecolor{splashedwhite}{rgb}{1.0, 0.99, 1.0}
\definecolor{lavendergray}{rgb}{0.77, 0.76, 0.82}
\definecolor{lightgray}{rgb}{0.83, 0.83, 0.83}
\definecolor{lavendermist}{rgb}{0.9, 0.9, 0.98}
\definecolor{lightgreen}{HTML}{f8fcf4}
\definecolor{lightblue}{HTML}{dfebf7}
\definecolor{zeroshot}{rgb}{0.9, 0.9, 0.9}
\definecolor{fourshot}{rgb}{0.8, 0.9, 0.8}
\definecolor{eightshot}{rgb}{0.8, 0.8, 0.9}
\definecolor{sixteenshot}{rgb}{0.9, 0.8, 0.8}
\definecolor{blue-violet}{rgb}{0.54, 0.17, 0.89}
\definecolor{coral}{HTML}{FF7F50}

\newcommand{\bench}{MMSI-Video-Bench} 

\def\paperID{12530} 
\def\confName{CVPR}
\def\confYear{2026}

\title{\bench: A Holistic Benchmark for Video-Based Spatial Intelligence}

\author{%
\textbf{Jingli Lin}$^{1,2*}$\quad \textbf{Runsen Xu}$^{1,3*\dagger}$\quad \textbf{Shaohao Zhu}$^{1,4}$\quad \textbf{Sihan Yang}$^{1}$ \quad \textbf{Peizhou Cao}$^{1,5}$\\ 
\textbf{Yunlong Ran}$^{1,4}$\quad \textbf{Miao Hu}$^{6}$\quad
\textbf{Chenming Zhu}$^{1,7}$\quad  \textbf{Yiman Xie}$^{1,4}$\quad 
\textbf{Yilin Long}$^{1,8}$\quad \textbf{Wenbo Hu}$^{1,9}$\\ 
\textbf{Dahua Lin}$^{1,3}$\quad \textbf{Tai Wang}$^{1 \textrm{\Letter}}$\quad \textbf{Jiangmiao Pang}$^{1 \textrm{\Letter}}$
\\[4pt]
$^{1}$Shanghai AI Laboratory\quad
$^{2}$Shanghai Jiaotong University \quad
$^{3}$The Chinese University \\ of Hong Kong \quad
$^{4}$Zhejiang University \quad
$^{5}$Beihang University\quad
$^{6}$Xi'an Jiaotong University \\
$^{7}$University of Hong Kong\quad
$^{8}$Fudan University \quad
$^{9}$University of California, Los Angeles\\[6pt]
\textbf{$^{*}$ Equal Contribution}\quad  \textbf{$^{\dagger}$ Project Lead}\\[6pt]
\\
{\worldwideweb \href{https://rbler1234.github.io/MMSI-VIdeo-Bench.github.io/}{{\text{Project Page}}}} \quad \quad {\github \href{https://github.com/InternRobotics/MMSI-Video-Bench}{{\text{Evaluation Code}}}}
\quad \quad
{\huggingface \href{https://huggingface.co/datasets/rbler/MMSI-Video-Bench}{{\text{MMSI-Video-Bench}}}}
}

\begin{document}

\twocolumn[{%
\renewcommand\twocolumn[1][]{#1}%

\maketitle
\begin{center}
    \centering
    \vspace{-1em}
    \captionsetup{type=figure}
   \includegraphics[width=0.98\linewidth]{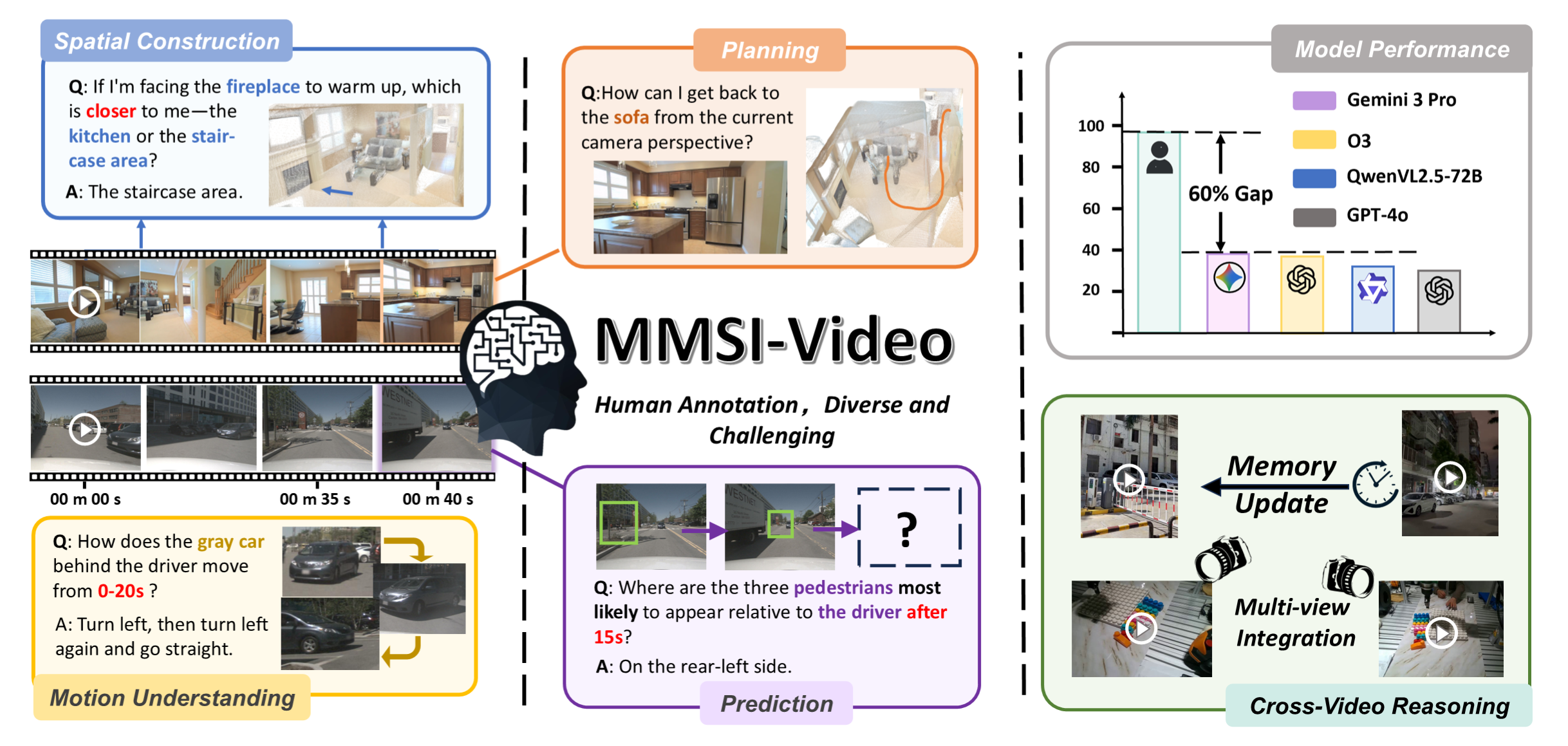}
   \vspace{-0.5em}
    \captionof{figure}{\textbf{\bench} is a diverse, human-annotated, and challenging benchmark, designed to evaluate models’ video-based spatial intelligence, including their ability to perceive, understand, reason, and make decisions over spatio-temporal information in videos. The bar chart in the top-right corner illustrates the substantial performance gap between state-of-the-art models and human performance.}
    \label{fig:teaser}
\end{center}
}]

\begin{abstract}
Spatial understanding over continuous visual input is crucial for MLLMs to evolve into general-purpose assistants in physical environments. Yet there is still no comprehensive benchmark that holistically assesses the progress toward this goal. In this work, we introduce \textbf{MMSI-Video-Bench}, a fully human-annotated benchmark for video-based spatial intelligence in MLLMs. It operationalizes a four-level framework, Perception, Planning, Prediction, and Cross-Video Reasoning, through 1{,}106 questions grounded in 1{,}278 clips from 25 datasets and in-house videos. Each item is carefully designed and reviewed by 3DV experts with explanatory rationales to ensure precise, unambiguous grounding. Leveraging its diverse data sources and holistic task coverage, MMSI-Video-Bench also supports three domain-oriented sub-benchmarks (Indoor Scene Perception Bench, Robot Bench and Grounding Bench) for targeted capability assessment. We evaluate 25 strong open-source and proprietary MLLMs, revealing a striking human--AI gap: many models perform near chance, and the best reasoning model lags humans by nearly 60\%. We further find that spatially fine-tuned models still fail to generalize effectively on our benchmark. Fine-grained error analysis exposes systematic failures in geometric reasoning, motion grounding, long-horizon prediction, and cross-video correspondence. We also show that typical frame-sampling strategies transfer poorly to our reasoning-intensive benchmark, and that neither 3D spatial cues nor chain-of-thought prompting yields meaningful gains. We expect our benchmark to establish a solid testbed for advancing video-based spatial intelligence.
\end{abstract}    
\section{Introduction}

For decades, humans have aspired to build an embodied general-purpose AI assistant, akin to JARVIS in \emph{Iron Man}. As MLLMs~\cite{bai2025qwen25vltechnicalreport, zhu2025internvl3exploringadvancedtraining, zhang2025llavavideovideoinstructiontuning, xu2025multi, chen2025expandingperformanceboundariesopensource} have begun to exhibit strong language and visual intelligence, they are increasingly viewed as a promising foundation for embodied AGI. One of the most important remaining challenges in this pursuit is to endow MLLMs with spatial intelligence, that is, the ability to perceive, reason about, and interact with physical space from continuous visual inputs, as humans do.

To reliably measure progress on this research, we need rigorous benchmarks. However, existing benchmarks have significant limitations: most operate on discrete single images~\cite{erqa, chen2024spatialvlmendowingvisionlanguagemodels, spatialrgpt} or multiple images~\cite{blink, yang2025mmsi, omnispatial25} rather than videos, leaving an input gap to the practical setting. Recent video-based benchmarks~\cite{yang2024think, zhang2025from,li2025sti,lin2025ostbench} also suffer from the following issues: (1) question types are not sufficiently holistic; (2) they rely heavily on templated automatic question generation, which restricts question diversity and may introduce template overfitting or biases~\cite{yang2025mmsi}; and (3) data sources and scenes are not comprehensive enough.

In this work, we introduce \textbf{MMSI-Video-Bench} to fill these gaps. We build our benchmark around a holistic, multi-level framework for video-based spatial intelligence consisting of Perception, Planning, Prediction, and Cross-Video Reasoning (see \Cref{fig:teaser}). Models are required to reason over a single video for spatial perception, capturing global scene information (Spatial Construction) and ego-/exo Motion dynamics. Beyond perception, models should be able to make decisions or take actions to interact with the environment (Planning) and further make Predictions about future spatial states of the world. Finally, a general spatial intelligence model should be capable of Cross-Video Reasoning for multi-view integration and memory updating.

We instantiate the above holistic framework as a diverse, accurate, and challenging MCQ benchmark. We adopt a fully human-designed 
protocol following~\cite{yang2025mmsi}. 
Eleven 3DV researchers manually design each sample, including selecting a video clip from a curated pool of about 20K videos, designing a novel question, writing the correct answer, distractors, and a brief rationale. A multi-stage review process leverages these rationales to ensure accuracy and unambiguity. Our video pool combines 25 open-source datasets with newly recorded in-house videos, covering a wide range of scenarios, including indoor scans, outdoor driving, robotics, etc. In total, with 400+ hours of annotation and verification, we obtain 1{,}106 questions grounded in 1{,}278 video clips, grouped into five task categories and 13 subtypes. Thanks to the diversity of data sources and the holistic coverage of task types in MMSI-Video-Bench, we are also able to build three domain-oriented sub-benchmarks: Indoor Scene Perception Bench, Robot Bench, and Grounding Bench, enabling targeted assessment of specific model capabilities.

Using MMSI-Video-Bench, we conduct a comprehensive evaluation of open-source and proprietary MLLMs. Current models remain far from the desired level of video-based spatial intelligence: many perform close to random guessing, and the best model, Google's Gemini 3 Pro, still trails humans by nearly 60\%. To the best of our knowledge, our benchmark yields the largest human--AI performance gap among existing video-based spatial benchmarks. In addition, we also evaluate models that have been spatially fine-tuned, yet their capabilities still fail to generalize effectively on our benchmark, further underscoring the challenge posed by MMSI-Video-Bench.

To provide diagnostic signals for future research, we perform per-category error analysis. For Spatial Construction, failures are dominated by geometric reasoning errors; for Motion, by fine-grained grounding failures on fast, subtle, or long-duration motion; for Planning and Prediction, by prompt--evidence misalignment where models ignore video cues; and for Cross-Video Reasoning, by difficult grounding and matching correspondences across videos. Beyond analysis, our preliminary exploration shows that neither 3D spatial cues nor chain-of-thought prompting yields meaningful gains on MMSI-Video-Bench. We further observe that, due to the reasoning-intensive nature of our benchmark, the commonly used frame-sampling strategy AKS~\cite{tang2025adaptivekeyframesamplinglong} does not transfer directly to {\bench}, leading to additional performance degradation. Overall, current MLLMs still have substantial room for improvement in spatial intelligence, and MMSI-Video-Bench offers an accurate and challenging testbed with actionable insights for advancing video-based spatial reasoning.



\begin{table*}[ht]
\centering
\scalebox{0.7}{
\setlength{\tabcolsep}{10pt}
\renewcommand{\arraystretch}{1.1}
\begin{tabular}{@{}ccccccc@{}}
\toprule
\textbf{Benchmark} &  \textbf{Source Diversity}                  & \textbf{Modality} & \textbf{Annotation Method} & \textbf{Task}                                  & \textbf{Samples Num} & \textbf{Human-AI Gap}      \\
\midrule

SpatialRGPT~\cite{spatialrgpt}      & -                                      & Single-Image        & Auto              & local-SU.                             & 1,406             & \textless{}42    \\
SpatialVLM~\cite{chen2024spatialvlmendowingvisionlanguagemodels}       & -                                      & Single-Image        & Auto\&Human       & local-SU.                             & 546               & \textless{}30    \\
CVBench~\cite{cvbench2025}          & 3               & Single-Image        & Auto              & local-SU.                             & 2,638             &  -                \\
MultiSPA~\cite{xu2025multi} & 3 & Multi-Image & Auto & local-SU. \& short-MU. & 7800 & - \\
All-Angles-Bench~\cite{yeh2025seeing} & 2 & Multi-Image & Human & SU. & 2100 & 21.2 \\
MMSI-Bench~\cite{yang2025mmsi}       & 8                                      & Multi-Image         & Human             & local-SU. \& short-MU.                & 1,000             & 55.3             \\
VSI-Bench~\cite{yang2024think}        & 3     & Video               & Auto              & SU. \& Plan.                        & 5,000             & 33               \\
OST-Bench~\cite{lin2025ostbench}       & 3  & Video               & Auto              & SU. \& CAM.-MU.                        & 10,000            & 29.3             \\
SPAR-Bench~\cite{zhang2025from}       & 3 & Video               & Auto              & SU. \& CAM.-MU.                        & 7207              & 27.8             \\
STI-Bench~\cite{li2025sti}        & 3           & Video               & Auto              & SU. \& INST./CAM.-MU.                            & 2,064             & -                \\
EgoExoBench~\cite{he2025egoexobenchbenchmarkfirstthirdperson}      & 6                                      & Multi-Video         & Auto\&Human       & CV.                                   & 7,330             & 41.9             \\

\midrule
MMSI-Video (ours) & 26                        & Video/Multi-Video   & Human             & SU. \& MU. \& Plan./Pred. \& CV. & 1,106             & 58.4 \\
\bottomrule
\end{tabular}
}
\vspace{-0.6em}
\caption{Comparison of MMSI-Video-Bench with other spatial reasoning benchmarks, highlighting its diversity, comprehensiveness, and challenge. INST. and CAM. refer to instance and camera, SU., MU., and CV. refer to Spatial Understanding, Motion Understanding, and Cross-Video Reasoning, respectively. The Human–AI Gap indicates the performance difference as a percentage.}
\label{tab:camprison_bench}
\end{table*}
\section{Related Work}
\textbf{Video-based Benchmarks.} Video-based benchmarks evaluate a model’s ability to perceive, understand, and reason over temporal and visual information. Early video benchmarks such as MSVD-QA, MSRVTT-QA~\cite{msvd-qa}, and ActivityNet-QA~\cite{activitynet-qa} mainly assess global visual comprehension, with limited attention to temporal understanding. Later works like NeXT-QA~\cite{nextqa} and MVBench~\cite{mvbench} emphasize temporal dynamics. Subsequently, benchmarks such as LongVideoBench~\cite{longvideobench} and Video-MME~\cite{videomme} further extended evaluation beyond surface-level perception, incorporating temporal–event reasoning. With the advancement of MLLMs, more video-based reasoning benchmarks have emerged, each designed to probe specific aspects of real-world understanding. These include benchmarks focusing on such as complex spatial reasoning within videos~\cite{li2025sti,yang2024think,zhang2025from,lin2025ostbench}, online inference under continuous observation~\cite{lin2025ostbench,li2025ovobenchfarvideollmsrealworld}, and cross-video reasoning~\cite{he2025egoexobenchbenchmarkfirstthirdperson}. Different from previous benchmarks, our proposed {\bench} serves as a more holistic and challenging benchmark for video spatial intelligence, covering complex reasoning about spatial layouts, motion understanding, and decision-making, as well as reasoning across multiple videos.

\noindent\textbf{Spatial Intelligence Benchmarks.} 
Existing benchmarks for evaluating spatial intelligence in multimodal large language models (MLLMs) vary substantially in task design, modality, scene scope, and the specific spatial abilities they aim to assess.
Early benchmarks such as SpatialRGPT~\cite{spatialrgpt}, SpatialVLM~\cite{chen2024spatialvlmendowingvisionlanguagemodels}, and CVBench~\cite{cvbench2025} focus on single-image spatial reasoning, emphasizing depth and distance. Later, video-based benchmarks, VSI-Bench~\cite{yang2024think}, SPAR-Bench~\cite{zhang2025from}, and OST-Bench~\cite{lin2025ostbench}, extend this to indoor scenes with object–object and object–camera relations, yet remain constrained by limited scene diversity and static spatial contexts.
More recent benchmarks, including MultiSPA~\cite{xu2025multi}, MMSI-Bench~\cite{yang2025mmsi}, and STI-Bench~\cite{li2025sti}, incorporate dynamic environments and cover a wider range of indoor–outdoor scenarios. MultiSPA and MMSI-Bench serve as demanding benchmarks focusing on multi-image local spatial localization and short-term motion understanding, whereas STI-Bench targets numerical estimation of instance-centric spatial states and motion trajectories within videos. 
Nevertheless, existing benchmarks focus on limited aspects or scene types, lacking a holistic evaluation across diverse real-world contexts. In contrast, our {\bench}, curated from diverse real-world videos and fully human-annotated, offers a more holistic and realistic assessment of spatial intelligence in MLLMs.

\section{\bench}

In this section, we introduce the formulation of our {\bench} and the methodology for its construction.

\begin{figure*}
  \centering
  \includegraphics[width=1.0\linewidth]{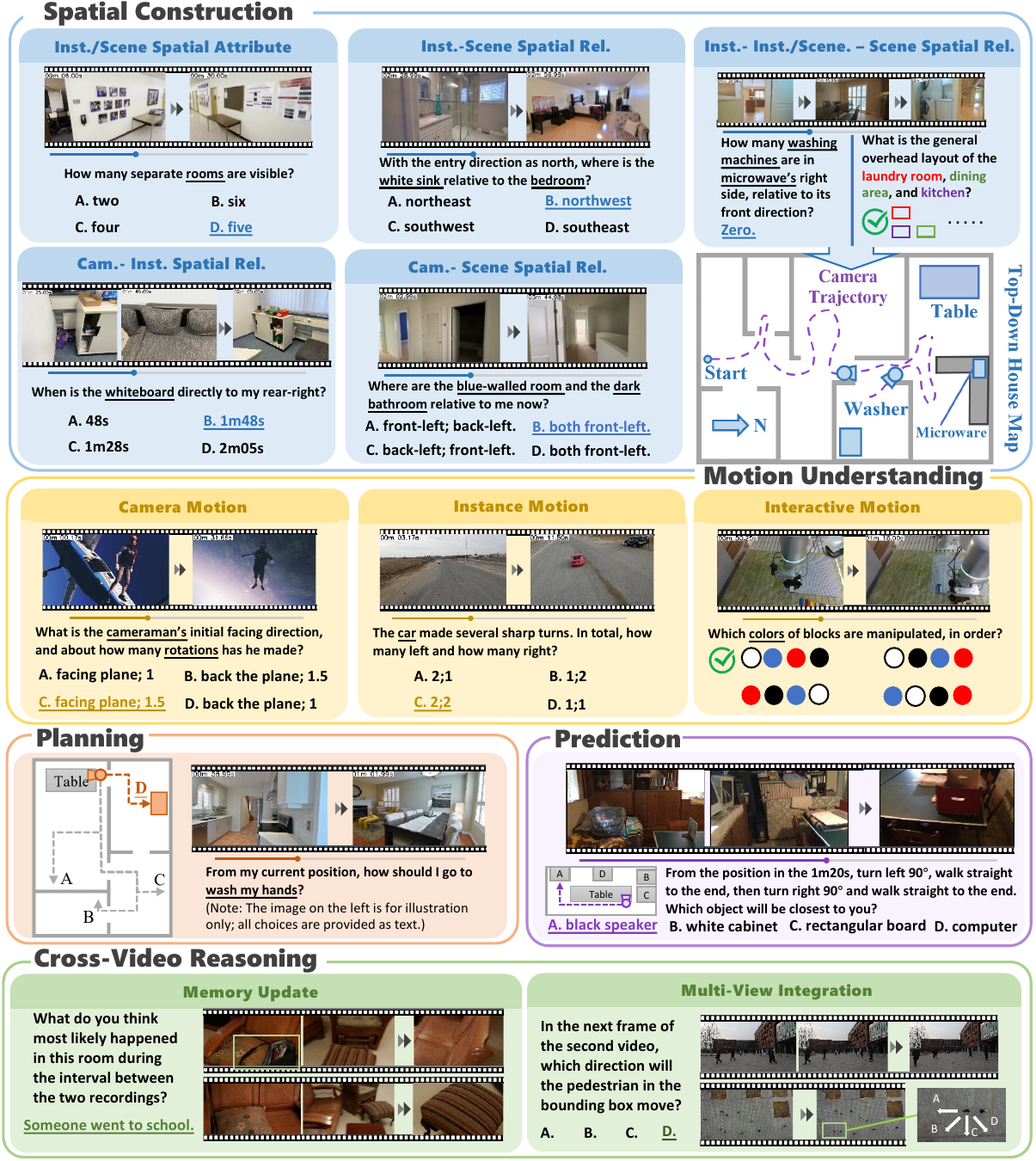}
  \vspace{-3ex}
\caption{Illustrative examples of different subtypes in MMSI-Video-Bench. Rel., Inst., and Cam. stand for Relationship, Instance, and Camera. Please refer to our project page for the full demo.}
  \vspace{-3ex}
  \label{fig:benchmark samples}
\end{figure*}

\subsection{Overview}

As a video-based spatial intelligence benchmark, {\bench} primarily evaluates a model’s capability to perceive, understand, and reason over video information, encompassing two core dimensions:

\noindent\textbf{Spatial.} This dimension concerns the spatial states (e.g., position, shape) of entities such as instances, scenes, or cameras, and their spatial relations (e.g., front–back, left–right, near–far) at a fixed moment. To assess this ability, we introduce the \textit{Spatial Construction} category, which requires the model to infer fine-grained global spatial layouts from partial and sequential video observations.

\noindent\textbf{Spatio-Temporal.} When motion occurs in the video (from the camera or instances), the spatial configuration changes over time. The \textit{Motion Understanding} category evaluates the model’s ability to reason about long-term motion dynamics across consecutive frames, including camera motion, individual instance motion, and interactive motion arising from interactions among multiple instances.

After understanding spatio-temporal information, the next step is decision-making based on video understanding. The \textit{Planning} and \textit{Prediction} categories focus on higher-level reasoning over sparse video information:(1) Planning tasks require the model to devise actions toward a specific goal based on visual cues; (2) Prediction tasks test the model’s ability to infer or imagine outcomes under hypothetical conditions.

The above categories comprehensively cover spatial intelligence within a single video.
 However, real-world understanding often requires reasoning across multiple videos.
 To achieve a more holistic evaluation, we extend {\bench} with \textit{Cross-Video Reasoning}, including:
 
 (1) Memory Update. From a temporal perspective, observations of the same scene in real-world settings are often temporally discontinuous (i.e., we do not continuously stay in one place). The model must therefore retain contextual memory from past observations and update it as new information becomes available.

(2) Multi-View Integration. From a spatial perspective, a single viewpoint rarely captures the complete spatial–temporal information of a complex scene. The model must thus integrate observations from multiple viewpoints to construct a unified representation of the scene.

Fig.\ref{fig:benchmark samples} illustrates example cases for each subtype. In the supplementary material, we provide an overview table that describes the details of each subtype.

\begin{figure*}
  \centering
  \includegraphics[width=1.0\linewidth]{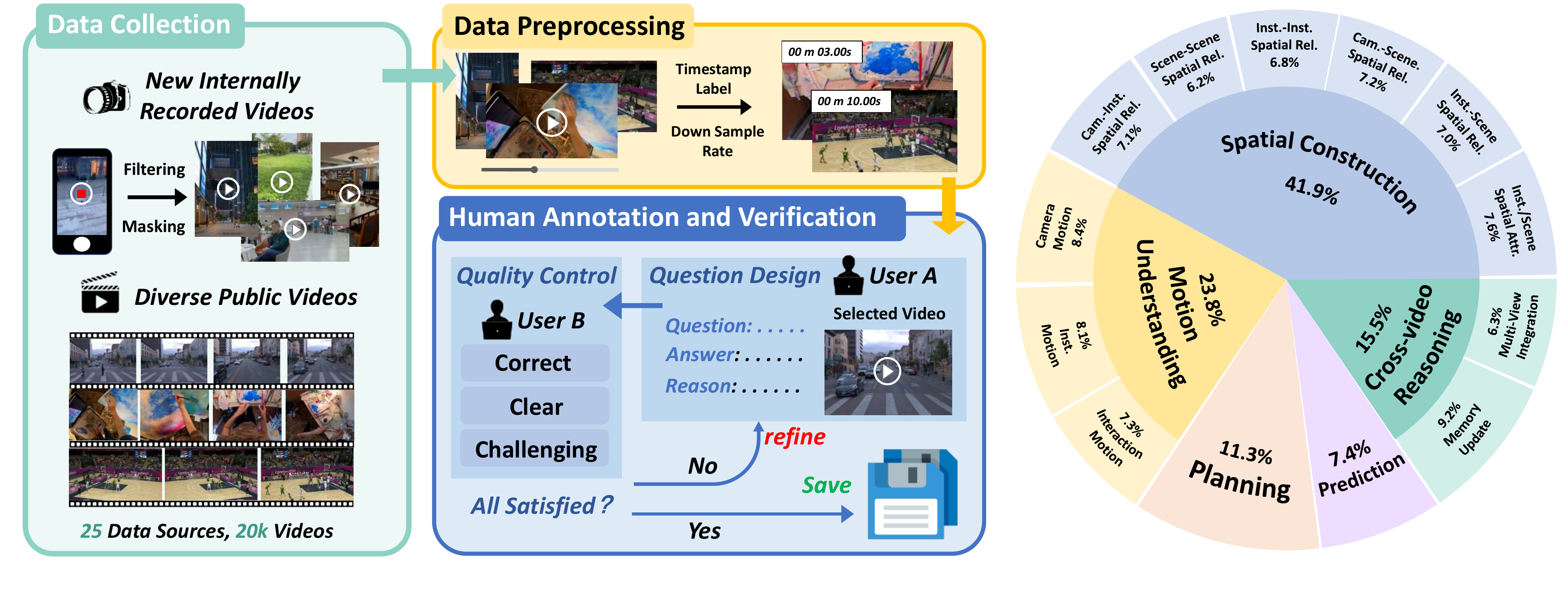}
  \vspace{-2.8em}
  \caption{(Left) The pipeline of benchmark sample construction. (Right) The distribution of question types in {\bench}.}
  \vspace{-3ex}
  \label{fig:data pipeline}
\end{figure*}
\subsection{Benchmark Construction}

\textbf{Data Collection \& Preprocessing.}
To ensure diversity in our benchmark, we curated videos spanning a broad spectrum of real-world scenarios. The collection includes a wide range of capture types, such as tabletop recordings, indoor scenes from single-room to multi-floor environments, outdoor building and street views, natural landscapes, sports activities, and movie footage. Our data sources include 25 publicly available datasets~\cite{roomtour3d,scannet,scannet++,3rscan,arkitscenes,jensen2014large,LaSOT,uav123,müller2018trackingnetlargescaledatasetbenchmark,nuScenes,EPICKITCHENS}, such as ScanNet~\cite{scannet}, RealEstate10k~\cite{realestate}, DL3DV~\cite{dl3dv}, Ego4D~\cite{ego4d}, DROID~\cite{khazatsky2024droid}, Waymo~\cite{Waymo} and MultiSports~\cite{Li_2021_ICCV} (the remaining datasets are listed in the supplementary material), amounting to approximately 20k video clips. In addition, we manually recorded and carefully selected 140 supplemental in-house videos, each anonymized via masking to protect personal privacy.
All videos were then downsampled to an appropriate frame rate for each category such that no key information would be lost. Each frame was timestamped in the top-left corner (formatted as “xx min yy s”) to facilitate precise temporal referencing during question design.

\noindent\textbf{Human Annotation.}
As shown in Fig.\ref{fig:data pipeline}, all annotations were conducted by a team of eleven trained researchers specializing in 3D vision. To preserve data diversity, annotation tasks were assigned so that each annotator received a balanced mix of task categories. We developed a dedicated annotation interface that allowed annotators to view all videos and design questions directly within the interface. Based on the provided visual context, annotators composed questions, corresponding answers, and concise reasoning, particularly for challenging items, to facilitate subsequent verification. All questions were designed as multiple-choice items, containing between four and six answer options and optionally adopting an interleaved image–text format.

\noindent\textbf{Data Quality Control.}
We implemented a strict acceptance protocol, where eleven researchers performed cross-evaluation, reviewing each other's work. Each annotation had to meet three criteria: clear (the question is unambiguous and clearly stated), correctness (the answer is unique and factually accurate), and challenge (the question requires non-trivial reasoning). Only annotations that met all three criteria were accepted, with a 100\% approval rate required. Annotations that failed were revised based on feedback and resubmitted for reevaluation.

\begin{figure}
  \centering
  \includegraphics[width=1.0\linewidth]{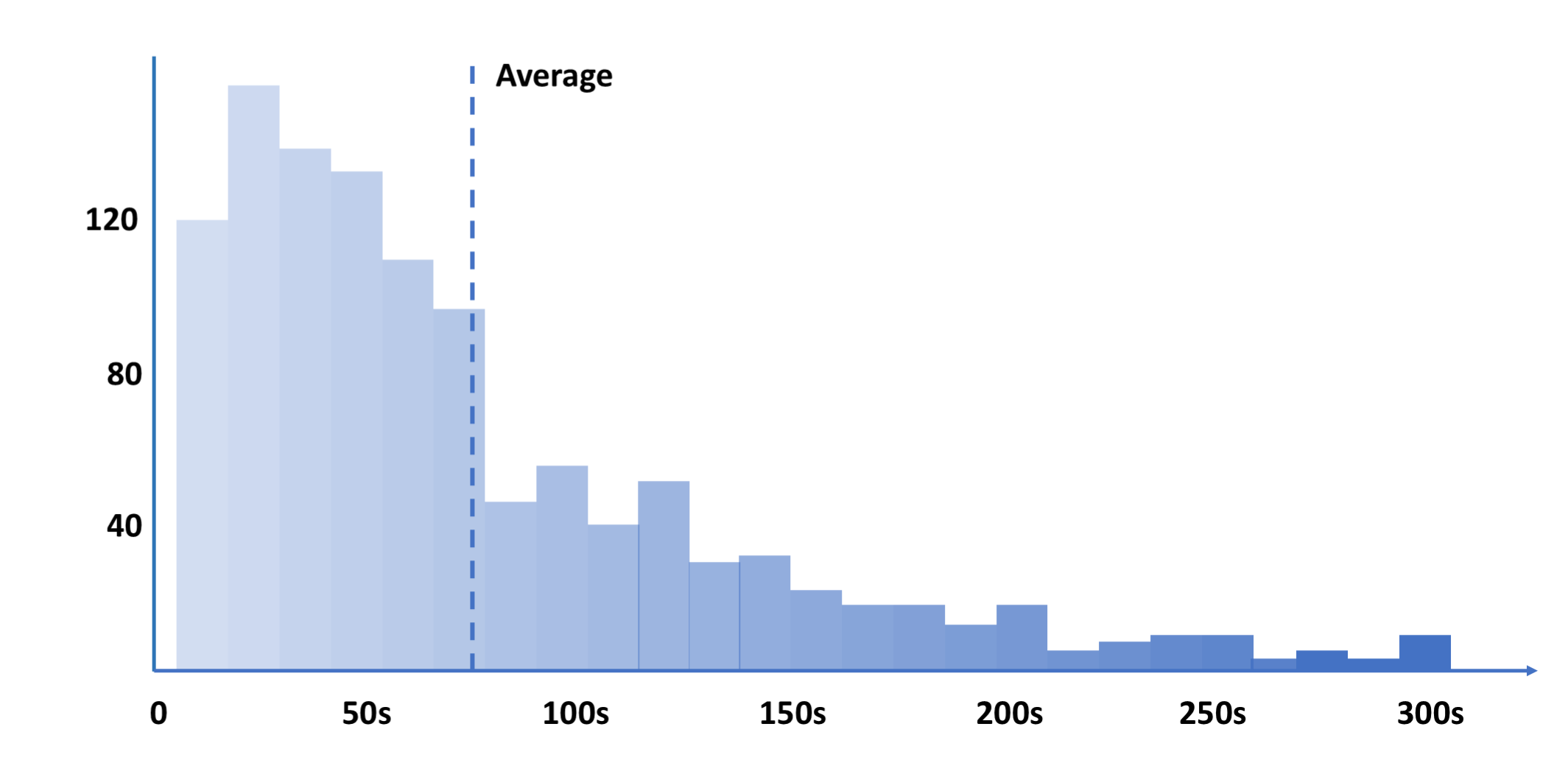}
  \vspace{-3ex}
  \caption{Duration distribution of all video samples in MMSI-Video-Bench (in seconds).}
  \vspace{-3ex}
  \label{fig:dis1}
\end{figure}

\noindent\textbf{Statistics.}
After the construction of benchmark samples, the final {\bench} contains 1,106 questions grounded in 1,278 video clips, covering five main categories and 13 subtypes, with their distribution shown in Fig.\ref{fig:data pipeline}. The average video duration is 1 minute 12 seconds, and the average question length is 164.5 characters. In Fig.\ref{fig:dis1}, we present the distribution of video durations.

\begin{table*}[tbp!]
  \centering
  \scriptsize
  \setlength{\tabcolsep}{1.3pt}
  \begin{adjustbox}{width=\textwidth,center}
    \begin{tabularx}{\textwidth}{%
      l
      *{6}{>{\centering\arraybackslash}p{0.065\textwidth}}  
      *{3}{>{\centering\arraybackslash}p{0.050\textwidth}}  
      *{2}{>{\centering\arraybackslash}p{0.050\textwidth}}  
      >{\centering\arraybackslash}p{0.050\textwidth}       
      >{\centering\arraybackslash}p{0.050\textwidth} 
      >{\centering\arraybackslash}p{0.050\textwidth}       
    }
    \toprule
    \multirow{2}{*}{\textbf{Models}}
      & \multicolumn{6}{c}{\cellcolor{cyan!8}\textbf{Spatial Construction}}
      & \multicolumn{3}{c}{\cellcolor{yellow!30}\textbf{Motion Understanding}}
      & \multicolumn{2}{c}{\cellcolor{green!15}\textbf{Cross-Video}}
 
      & \cellcolor{orange!20}\makecell{\textbf{Plan.}}
      & \cellcolor{purple!25}\makecell{\textbf{Pred.}}
      & \multirow{2}{*}{\textbf{Avg.}} \\
    & \scriptsize Attr. & \scriptsize Inst.-Inst. & \scriptsize Inst.–Scen.
    & \scriptsize Scen.–Scen. & \scriptsize Cam.-Inst. & \scriptsize Cam.–Scen.
    & \scriptsize Cam. & \scriptsize Inst. & \scriptsize Inter.
    & \scriptsize MU. & \scriptsize MV. & -
    & - & \\
    \midrule
    \rowcolor{gray!10}\multicolumn{15}{l}{\emph{\textbf{(Sufficient-Coverage)  Proprietary}}} \\
O4-mini &31.3 &35.5 &39.0 &30.4 &36.7 &36.2 &\textbf{45.2} &28.9 &30.9 &40.2 &31.4 &\textbf{38.7} &26.8 &35.1\\
O3 &39.8 &31.6 &\textbf{46.8} &\textbf{36.2} &\textbf{43.0} &36.2 &34.4 &\textbf{34.4} &38.3 &\textbf{44.1} &34.3 &34.7 &\textbf{31.7} &\cellcolor{green!60}{\textbf{37.3}}\\
GPT-4o &22.9 &30.3 &35.1 &26.1 &26.6 &27.5 &24.7 &28.9 &24.7 &34.3 &25.7 &29.8 &26.8 &28.1\\
Gemini 2.5 Flash &\textbf{47.0} &36.8 &40.3 &31.9 &30.4 &\textbf{48.8} &38.7 &30.0 &\textbf{43.2} &37.2 &30.0 &32.3 &30.5 &36.6\\
-Thinking &43.4 &\textbf{38.2} &41.6 &24.6 &34.2 &36.2 &44.1 &\textbf{34.4} &35.8 &36.3 &\textbf{38.6} &36.3 &\textbf{31.7} &\cellcolor{green!20}{36.7}\\

    \midrule
    \rowcolor{gray!10}\multicolumn{15}{l}{\emph{\textbf{(Sufficient-Coverage)  Open-source}}} \\
InternVL2.5-8B &26.5 &27.6 &26.0 &\textbf{36.2} &19.0 &30.0 &29.0 &33.3 &30.9 &34.3 &22.9 &24.2 &32.9 &28.7\\
InternVL3-8B &28.9 &\textbf{38.2} &27.3 &31.9 &22.8 &23.8 &36.6 &31.1 &30.9 &30.4 &34.3 &26.6 &23.2 &29.6\\
InternVideo2.5-8B &26.5 &26.3 &23.4 &27.5 &22.8 &28.8 &26.9 &23.3 &25.9 &34.3 &21.4 &29.0 &29.3 &26.9\\
QwenVL2.5-7B &31.3 &21.1 &19.5 &30.4 &19.0 &31.2 &35.5 &\textbf{36.7} &29.6 &27.4 &30.0 &26.6 &\textbf{35.4} &28.8\\
QwenVL2.5-32B &\textbf{33.7} &28.9 &31.2 &31.9 &\textbf{30.4} &30.0 &\textbf{43.0} &35.6 &25.9 &25.5 &\textbf{41.4} &33.1 &30.5 &\cellcolor{green!60}{\textbf{32.4}}\\
QwenVL2.5-72B &\textbf{33.7} &22.4 &28.6 &30.4 &24.1 &28.8 &32.3 &33.3 &39.5 &\textbf{35.3} &35.7 &\textbf{35.5} &30.5 &\cellcolor{green!20}{31.8}\\
QwenVL3-8B &27.8 &31.5 &\textbf{33.3} &26.9 &23.1 &22.4 &27.3 &29.6 &\textbf{41.0} &28.4 &29.9 &26.5 &32.0 &29.1\\
QwenVL3-30B &31.3 &28.9 &26.0 &24.6 &22.8 &\textbf{36.2} &29.0 &21.1 &33.3 &29.4 &28.6 &30.6 &\textbf{35.4} &29.1\\
-Thinking &27.7 &36.8 &29.9 &33.3 &\textbf{30.4} &25.0 &26.9 &23.3 &30.9 &25.5 &25.7 &26.6 &25.6 &28.0\\
    \midrule
    \rowcolor{gray!10}\multicolumn{15}{l}{\emph{\textbf{(Uniform-50)  Proprietary}}} \\
Claude-haiku-4.5 &31.3 &31.6 &32.5 &34.8 &32.9 &35.0 &35.5 &\textbf{41.1} &35.8 &33.3 &\textbf{38.6} &31.4 &32.9 &34.3\\
GPT-5 &44.6 &38.2 &\textbf{44.2} &\textbf{39.1} &40.5 &41.2 &37.6 &28.9 &\textbf{42.0} &33.3 &28.6 &34.7 &28.1 &36.8\\
O4-mini &36.1 &34.2 &36.4 &33.3 &39.2 &42.5 &37.6 &32.2 &37.0 &32.4 &30.0 &28.2 &28.1 &34.2\\
O3 &34.9 &34.2 &40.3 &36.2 &40.5 &43.8 &37.6 &37.8 &35.8 &\textbf{41.2} &25.7 &\textbf{41.1} &26.8 &\cellcolor{green!20}{37.0}\\
GPT-4o &26.5 &26.3 &31.2 &26.1 &38.0 &38.8 &36.6 &35.6 &27.2 &30.4 &28.6 &33.1 &29.3 &31.6\\
Gemini 3 Pro &44.6 &\textbf{39.5} &\textbf{44.2} &33.3 &29.1 &43.8 &35.5 &40.0 &38.3 &34.3 &37.1 &38.7 &\textbf{35.4} &\cellcolor{green!60}{\textbf{38.0}}\\
Gemini 2.5 Flash &37.4 &38.2 &40.3 &30.4 &\textbf{43.0} &43.8 &39.8 &31.1 &38.3 &30.4 &37.1 &30.6 &24.4 &35.4\\
-Thinking &43.4 &25.0 &42.9 &23.2 &32.9 &\textbf{47.5} &38.7 &28.9 &38.3 &31.4 &28.6 &40.3 &31.7 &35.2\\
Doubao-1.5-thinking &\textbf{45.1} &26.7 &18.8 &38.6 &29.3 &32.7 &33.3 &32.2 &40.0 &31.4 &30.8 &32.0 &20.4 &31.6\\
Seed-1.6-vision &43.1 &26.7 &16.7 &38.6 &34.5 &40.0 &\textbf{40.9} &35.6 &40.0 &32.9 &38.5 &37.3 &25.9 &34.9\\

    \midrule
    \rowcolor{gray!10}\multicolumn{15}{l}{\emph{\textbf{(Uniform-50)  Open-source}}} \\
InternVL2.5-8B &21.7 &36.8 &31.2 &30.4 &17.7 &31.2 &32.3 &27.8 &30.9 &35.3 &28.6 &26.6 &28.1 &29.1\\
InternVL2.5-38B &27.7 &30.3 &26.0 &27.5 &11.4 &33.8 &35.5 &34.4 &34.6 &35.3 &28.6 &33.1 &\textbf{40.2} &31.0\\
InternVL2.5-78B &33.7 &36.8 &\textbf{37.7} &18.8 &29.1 &32.5 &28.0 &34.4 &30.9 &31.4 &28.6 &\textbf{35.5} &26.8 &31.4\\
InternVL3-8B &33.7 &32.9 &29.9 &27.5 &19.0 &18.8 &\textbf{36.6} &32.2 &37.0 &31.4 &32.9 &33.1 &26.8 &30.4\\
InternVL3-38B &33.7 &36.8 &31.2 &24.6 &16.5 &36.2 &30.1 &30.0 &27.2 &30.4 &27.1 &24.2 &28.1 &28.8\\
InternVL3-78B &\textbf{43.4} &31.6 &35.1 &23.2 &\textbf{30.4} &31.2 &34.4 &32.2 &38.3 &37.2 &34.3 &27.4 &24.4 &\cellcolor{green!20}{32.5}\\
InternVideo2.5-8B &27.7 &32.9 &22.1 &23.2 &22.8 &28.8 &25.8 &30.0 &25.9 &27.4 &34.3 &29.8 &24.4 &27.4\\
LLaVA-Video-7B &27.7 &28.9 &28.6 &30.4 &19.0 &25.0 &31.2 &35.6 &30.9 &36.3 &20.0 &21.0 &35.4 &28.5\\
LLaVA-Video-72B &36.1 &\textbf{39.5} &28.6 &21.7 &15.2 &23.8 &30.1 &\textbf{42.2} &37.0 &28.4 &32.9 &29.0 &29.3 &30.4\\
QwenVL2.5-7B &26.5 &25.0 &29.9 &\textbf{34.8} &20.2 &\textbf{37.5} &33.3 &31.1 &34.6 &24.5 &22.9 &34.7 &28.1 &29.7\\
QwenVL2.5-32B &25.3 &26.3 &31.2 &20.3 &21.5 &30.0 &33.3 &35.6 &25.9 &28.4 &32.9 &27.4 &31.7 &28.6\\
QwenVL2.5-72B &32.5 &18.4 &\textbf{37.7} &29.0 &29.1 &28.8 &34.4 &41.1 &\textbf{40.7} &36.3 &\textbf{41.4} &26.6 &30.5 &\cellcolor{green!60}{\textbf{32.7}}\\
QwenVL3-8B &33.7 &25.0 &24.7 &26.1 &24.1 &30.0 &35.5 &22.2 &29.6 &31.4 &20.0 &25.0 &29.3 &27.6\\
QwenVL3-30B &27.7 &31.6 &33.8 &20.3 &19.0 &26.2 &29.0 &28.9 &32.1 &31.4 &31.4 &29.0 &31.7 &28.8\\
-Thinking &27.7 &36.8 &\textbf{37.7} &31.9 &25.3 &31.2 &26.9 &26.7 &28.4 &\textbf{39.2} &30.0 &28.2 &31.7 &30.8\\

    \midrule
    \rowcolor{gray!10}\multicolumn{15}{l}{\emph{\textbf{Baseline}}} \\
    Random Guessing    &24.1 & 23.7 & 24.4 & 24.4 & 24.2 & 24.3 & 24.1 & 24.9 & 24.8 & 23.1 & 23.2 & 24.8 & 24.4 & 24.1 \\
    Human Level   &95.2 &94.8 &96.3 &96.0 &92.8 &94.9 &96.8 &95.6 &93.7 &94.4 &92.0 &95.1 &94.2 &96.4   \\

    \bottomrule
    \end{tabularx}
  \end{adjustbox}
  \vspace{-1.1em}
  \caption{Performance of various models under the Sufficient-Coverage and Uniform-50 settings. The highest and second-highest average scores across settings are highlighted in dark green and light green, respectively. Attr., Inst., Cam., Scen., and Inter. denote Attribute, Instance, Camera, Scene, and Interaction, respectively.  MU., and MV. represent Memory Update, and Multi-View Integration, respectively.}
  \label{tab:main_results}
  \vspace{-1em}
\end{table*}
\section{Experiments}

\subsection{Evaluation Settings}

We evaluate a wide range of state-of-the-art open-source and proprietary models on {\bench}, including GPT-5/ O3/ O4-mini/ GPT-4o, Gemini 3 Pro/ Gemini 2.5 Flash, Claude-4.5, Seed-1.6-Vision, Doubao-1.5-thinking, InternVL series, QwenVL series, LLaVA-Video series, and others. All proprietary models are tested via their official APIs, while all open-source models are evaluated using 8×A100 GPUs and follow their officially released inference configurations.
Due to (i) the limited number of images allowed per request by some proprietary APIs, and (ii) out-of-memory issues when running large open-source models on long videos, we establish two evaluation tracks:

\noindent \textbf{Uniform-50.} Each model receives exactly 50 uniformly sampled frames from the original video. This configuration aligns with the recommended number of input frames for most evaluated models.

\noindent \textbf{Sufficient-Coverage.} In this setting, each model receives the complete set of frames used during annotation, ensuring no visual information is omitted. 

For comparison, we additionally provide two baselines: random guessing and human performance. Since all questions in {\bench} are multiple-choice, we adopt an exact-match accuracy metric, where a prediction is considered correct only if it exactly matches the ground-truth.

Besides general-purpose models, we also evaluate several models that are finetuned on spatial reasoning data or equipped with latent spatial representations. 

\subsection{Main Results}

We report the performance of various models on our benchmark. Tab. \ref{tab:main_results} summarizes the results of all evaluated models across different question subtypes under two evaluation settings.
From the results in the table, we can draw the following key observations:

\noindent\textbf{Substantial gap between MLLMs’ and humans’ performance.} Model performance across all question types in {\bench} falls significantly short of human-level performance. Most models achieve low scores, some approaching the level of random guessing, and even the best-performing model, Gemini 3 Pro (38.0), lags behind humans (96.4) by nearly 60\%. This indicates current models are unable to handle these challenging tasks, highlighting the inherent difficulty of {\bench}. This observation motivates a deeper investigation into the underlying reasons for the models’ subpar performance on this benchmark.

\noindent\textbf{MLLMs exhibit weaknesses across all categories.} 
Previous spatial reasoning benchmarks have primarily focused on evaluating models’ Spatial Construction abilities across various scenarios and contexts, consistently showing poor performance in this aspect~\cite{yang2024think, yang2025mmsi, lin2025ostbench}.
 Our benchmark provides a more holistic evaluation, extending beyond Spatial Construction to assess other dimensions such as Motion Understanding, Planning, Prediction, and Cross-Video Reasoning. As shown in Tab.\ref{tab:main_results}, we observe that models also struggle considerably in these areas. In the subsequent error analysis section, we further investigate the specific capability bottlenecks underlying these weaknesses.

\noindent\textbf{Sufficient-Coverage does not outperform Uniform-50.}
 The Sufficient-Coverage setting is assumed to result in better performance than the Uniform-50 setting, as it provides the most visual information.
 However, we observe that most models show no significant improvement, and in many cases, even a performance drop under Sufficient-Coverage. Prior work~\cite{lessmore} has likewise shown that more input frames can introduce redundancy that hinders reasoning. To further enhance model performance, it is crucial to develop more effective strategies for key frame sampling. An in-depth analysis of the impact of sampling strategies on model performance is presented in the frame sampling study section.

\noindent\textbf{Comparison across models.} We observe that proprietary models consistently outperform open-source ones. Among open-source models, the best-performing ones, QwenVL2.5-72B (Uniform-50, 32.7) and QwenVL2.5-32B (Sufficient-Coverage, 32.4), still exhibit a noticeable gap compared to most proprietary models under the same settings. Within open-source models, the results broadly follow the trend that larger parameter scales lead to better performance. In contrast, enabling the thinking mode brings only marginal improvements, as shown by comparing Gemini 2.5 Flash and QwenVL3-30B with their thinking mode versions.

\noindent\textbf{Prediction is the most challenging main category, and Camera–Instance Spatial Relation is the most challenging subtype.} In Tab.\ref{tab:main_results}, performance on Prediction is generally lower than other main task categories (i.e., the average scores of Spatial Construction, Motion Understanding, Planning, and Cross-Video Reasoning). This is because these tasks require models to go beyond simply understanding the spatio-temporal information and instead make predictions based on specific conditions or physical priors. Among the various types of Spatial Construction subtasks, the Camera–Instance Spatial Relation subtype is the most challenging. This is due to its combination of ego-to-scene spatial reasoning and detailed grounding of instances within the video, making it the most difficult among all subtypes.

\noindent\textbf{Model performance across main categories and difficulty Levels.} We further computed the average scores of each model across the main categories, as shown in Tab.\ref{tab:sub_table}. Overall, Gemini 3 Pro achieves the highest performance, while GPT-5 demonstrates the strongest capabilities in spatial construction and reasoning tasks. On the other hand, Seed-1.6-vision excels at motion understanding and reasoning across multiple video segments. As a model with high reasoning and decision-making capacity, Gemini 3 Pro also leads in Prediction and Planning tasks compared to other models. Furthermore, we categorize our benchmark into three difficulty levels: easy, medium, and hard, based on the overall accuracy of all models on each question. Given our categorization criteria, it is natural that the scores follow the trend: hard $<$ medium $<$ easy. Examining model performance across these difficulty levels, we find that Gemini 3 Pro and Gemini 2.5 Flash are particularly effective at solving questions that most other models struggle with.

\begin{table*}[ht]
\centering
\tiny
    \resizebox{0.9\textwidth}{!}{
\begin{tabular}{lcccccccc}
\toprule
\multicolumn{1}{l|}{Methods}                     & \multicolumn{1}{c|}{Avg}    & SC.      & MU.   & Plan.\&Pred.  & \multicolumn{1}{c|}{CV.}    &  Hard    & Medium          & Easy          \\ \hline
\rowcolor[HTML]{ECF4FF} 
\textit{(Sufficient-Coverage) Proprietary}                            &                                                   &               &               &                                    &               &               &               &               \\
\multicolumn{1}{l|}{O4-mini}&\multicolumn{1}{l|}{35.1}&34.9&35.2&34.0&\multicolumn{1}{l|}{36.6}&16.2&30.0&\textbf{56.7}\\
\multicolumn{1}{l|}{O3}&\multicolumn{1}{l|}{\cellcolor{green!60}{\textbf{37.3}}}&39.0&35.6&33.5&\multicolumn{1}{l|}{\textbf{40.1}}&19.3&35.0&55.4\\
\multicolumn{1}{l|}{GPT-4o}&\multicolumn{1}{l|}{28.1}&28.0&26.1&28.6&\multicolumn{1}{l|}{30.8}&13.2&26.0&43.3\\
\multicolumn{1}{l|}{Gemini 2.5 Flash}&\multicolumn{1}{l|}{36.6}&\textbf{39.4}&37.1&31.6&\multicolumn{1}{l|}{34.3}&\textbf{21.7}&32.8&53.6\\
\multicolumn{1}{l|}{-Thinking}&\multicolumn{1}{l|}{36.7}&36.6&\textbf{38.3}&\textbf{34.5}&\multicolumn{1}{l|}{37.2}&21.1&\textbf{35.2}&51.7\\
\hline
\rowcolor[HTML]{ECF4FF} 
\textit{(Sufficient-Coverage) Open-source}                            &                                                   &               &               &                                    &               &               &               &               \\
\multicolumn{1}{l|}{InternVL2.5-8B}&\multicolumn{1}{l|}{28.7}&27.4&31.1&27.7&\multicolumn{1}{l|}{29.6}&13.2&24.5&46.4\\
\multicolumn{1}{l|}{InternVL3-8B}&\multicolumn{1}{l|}{29.6}&28.7&33.0&25.2&\multicolumn{1}{l|}{32.0}&\textbf{15.6}&26.0&45.4\\
\multicolumn{1}{l|}{InternVideo2.5-8B}&\multicolumn{1}{l|}{26.9}&25.9&25.4&29.1&\multicolumn{1}{l|}{29.1}&14.4&21.2&43.5\\
\multicolumn{1}{l|}{QwenVL2.5-7B}&\multicolumn{1}{l|}{28.8}&25.4&34.1&30.1&\multicolumn{1}{l|}{28.5}&13.8&26.5&44.3\\
\multicolumn{1}{l|}{QwenVL2.5-32B}&\multicolumn{1}{l|}{\cellcolor{green!60}{\textbf{32.4}}}&\textbf{31.0}&\textbf{35.2}&32.0&\multicolumn{1}{l|}{32.0}&13.8&\textbf{31.0}&49.9\\
\multicolumn{1}{l|}{QwenVL2.5-72B}&\multicolumn{1}{l|}{31.8}&28.0&34.9&\textbf{33.5}&\multicolumn{1}{l|}{\textbf{35.5}}&13.8&24.8&\textbf{54.9}\\
\multicolumn{1}{l|}{QwenVL3-8B}&\multicolumn{1}{l|}{29.1}&27.4&32.3&28.7&\multicolumn{1}{l|}{29.0}&12.8&22.0&51.0\\
\multicolumn{1}{l|}{QwenVL3-30B}&\multicolumn{1}{l|}{29.1}&28.4&27.6&32.5&\multicolumn{1}{l|}{29.1}&12.2&22.0&51.2\\
\multicolumn{1}{l|}{-Thinking}&\multicolumn{1}{l|}{28.0}&30.4&26.9&26.2&\multicolumn{1}{l|}{25.6}&11.9&23.0&47.2\\
\hline
\rowcolor[HTML]{ECF4FF} 
\textit{(Uniform-50) Proprietary}                            &                                                   &               &               &                                    &               &               &               &               \\
\multicolumn{1}{l|}{Claude-haiku-4.5}&\multicolumn{1}{l|}{34.3}&33.0&37.5&32.0&\multicolumn{1}{l|}{\textbf{35.5}}&18.0&30.5&52.2\\
\multicolumn{1}{l|}{GPT-5}&\multicolumn{1}{l|}{36.8}&\textbf{41.4}&36.0&32.0&\multicolumn{1}{l|}{31.4}&20.2&32.5&55.7\\
\multicolumn{1}{l|}{O4-mini}&\multicolumn{1}{l|}{34.2}&37.1&35.6&28.2&\multicolumn{1}{l|}{31.4}&15.0&29.0&\textbf{56.2}\\
\multicolumn{1}{l|}{O3}&\multicolumn{1}{l|}{37.0}&38.4&37.1&35.4&\multicolumn{1}{l|}{34.9}&20.5&34.2&54.1\\
\multicolumn{1}{l|}{GPT-4o}&\multicolumn{1}{l|}{31.6}&31.2&33.3&31.6&\multicolumn{1}{l|}{29.6}&17.7&29.2&45.9\\
\multicolumn{1}{l|}{Gemini 3 Pro}&\multicolumn{1}{l|}{\cellcolor{green!60}{\textbf{38.0}}}&39.2&37.9&\textbf{37.4}&\multicolumn{1}{l|}{\textbf{35.5}}&\textbf{22.9}&35.2&53.8\\
\multicolumn{1}{l|}{Gemini 2.5 Flash}&\multicolumn{1}{l|}{35.4}&39.0&36.4&28.2&\multicolumn{1}{l|}{33.1}&19.0&32.5&52.8\\
\multicolumn{1}{l|}{-Thinking}&\multicolumn{1}{l|}{35.2}&36.2&35.2&36.9&\multicolumn{1}{l|}{30.2}&19.3&33.0&51.2\\
\multicolumn{1}{l|}{Doubao-1.5-thinking}&\multicolumn{1}{l|}{31.6}&31.9&34.9&27.1&\multicolumn{1}{l|}{31.2}&13.2&31.9&48.1\\
\multicolumn{1}{l|}{Seed-1.6-vision}&\multicolumn{1}{l|}{34.9}&33.5&\textbf{38.9}&32.6&\multicolumn{1}{l|}{34.9}&16.0&\textbf{35.3}&51.5\\
\hline
\rowcolor[HTML]{ECF4FF} 
\textit{(Uniform-50) Open-source}                            &                                                   &               &               &                                    &               &               &               &               \\
\multicolumn{1}{l|}{InternVL2.5-8B}&\multicolumn{1}{l|}{29.1}&28.0&30.3&27.2&\multicolumn{1}{l|}{32.6}&13.8&23.2&48.5\\
\multicolumn{1}{l|}{InternVL2.5-38B}&\multicolumn{1}{l|}{31.0}&26.1&34.9&\textbf{35.9}&\multicolumn{1}{l|}{32.6}&11.9&27.5&51.2\\
\multicolumn{1}{l|}{InternVL2.5-78B}&\multicolumn{1}{l|}{31.4}&31.7&31.1&32.0&\multicolumn{1}{l|}{30.2}&14.7&25.0&52.5\\
\multicolumn{1}{l|}{InternVL3-8B}&\multicolumn{1}{l|}{30.4}&26.9&35.2&30.6&\multicolumn{1}{l|}{32.0}&13.8&27.5&47.8\\
\multicolumn{1}{l|}{InternVL3-38B}&\multicolumn{1}{l|}{28.8}&30.0&29.2&25.7&\multicolumn{1}{l|}{29.1}&9.2&28.0&46.7\\
\multicolumn{1}{l|}{InternVL3-78B}&\multicolumn{1}{l|}{32.5}&\textbf{32.8}&34.9&26.2&\multicolumn{1}{l|}{36.0}&14.4&27.8&53.3\\
\multicolumn{1}{l|}{InternVideo2.5-8B}&\multicolumn{1}{l|}{27.4}&26.3&27.3&27.7&\multicolumn{1}{l|}{30.2}&15.3&23.0&42.5\\
\multicolumn{1}{l|}{LLaVA-Video-7B}&\multicolumn{1}{l|}{28.5}&26.5&32.6&26.7&\multicolumn{1}{l|}{29.6}&\textbf{15.9}&27.0&40.9\\
\multicolumn{1}{l|}{LLaVA-Video-72B}&\multicolumn{1}{l|}{30.4}&27.6&36.4&29.1&\multicolumn{1}{l|}{30.2}&14.1&25.8&49.3\\
\multicolumn{1}{l|}{QwenVL2.5-7B}&\multicolumn{1}{l|}{29.7}&28.9&33.0&32.0&\multicolumn{1}{l|}{23.8}&11.3&\textbf{29.0}&46.2\\
\multicolumn{1}{l|}{QwenVL2.5-32B}&\multicolumn{1}{l|}{28.6}&25.9&31.8&29.1&\multicolumn{1}{l|}{30.2}&13.2&22.5&48.3\\
\multicolumn{1}{l|}{QwenVL2.5-72B}&\multicolumn{1}{l|}{\cellcolor{green!60}{\textbf{32.7}}}&29.3&\textbf{38.6}&28.2&\multicolumn{1}{l|}{\textbf{38.4}}&11.9&26.5&\textbf{57.3}\\
\multicolumn{1}{l|}{QwenVL3-8B}&\multicolumn{1}{l|}{27.6}&27.4&29.2&26.7&\multicolumn{1}{l|}{26.7}&8.0&21.8&50.7\\
\multicolumn{1}{l|}{QwenVL3-30B}&\multicolumn{1}{l|}{28.8}&26.5&29.9&30.1&\multicolumn{1}{l|}{31.4}&8.3&26.2&49.1\\
\multicolumn{1}{l|}{-Thinking}&\multicolumn{1}{l|}{30.8}&31.7&27.3&29.6&\multicolumn{1}{l|}{35.5}&11.9&27.5&50.7\\
\bottomrule

\end{tabular}
}

\vspace{-1ex}
\caption{Model performance across main categories and difficulty levels. "SC." abbrev for "Spatial Construction", "MU." abbrev for "Motion Understanding" and "CV." abbrev for "Cross-Video Reasoning".}

\label{tab:sub_table}
\vspace{-1ex}
\end{table*}

\subsection{Evaluation of Spatially Fine-tuned Models}
In recent years, several approaches have emerged to enhance general-purpose models with spatial reasoning capabilities, either by training them on spatial reasoning data (e.g., SpaceQwen~\cite{omnispatial25}) or by introducing architectural modifications to equip models with latent spatial representations (e.g., VLM3R~\cite{fan2025vlm3rvisionlanguagemodelsaugmented}, Spatial-MLLM~\cite{wu2025spatialmllmboostingmllmcapabilities}). These methods aim to endow models with spatial intelligence and have reported noticeable improvements on certain spatial reasoning benchmarks. We evaluate these spatially fine-tuned models under our benchmark as well under the Uniform-50 setting.

\begin{table*}[t]
\centering
\small
\begin{tabular}{l c cc cccc}
\toprule
Model  &  Modifications    & Avg   & SC.    & MU.    & Plan.\&Pred.    & Cross-Video    \\
\midrule
(base) LLaVA-Video-7B & -  & 28.48 & 26.51 & 32.58 & 26.70 & 29.65 \\
Spatial-MLLM  & Architecture    & 24.05{(\textcolor{red!100!black}{-4.43})} & 23.28 & 27.65 & 23.30  & 21.51 \\
VLM3R    & Architecture    & 4.97{(\textcolor{red!100!black}{-23.51})}  & 5.82  & 6.06  & 4.37  & 1.74  \\
\midrule
(base) QwenVL2.5-3B  & -  & 27.67 & 27.37 & 32.58 & 23.79 & 25.58 \\
SpaceQwen  & Fine-tuning  & 27.58{(\textcolor{red!100!black}{-0.09})} & 25.00 & 26.89 & 35.44 & 26.16 \\
\bottomrule
\end{tabular}

\caption{The table presents the performance of various spatially fine-tuned models and their corresponding base models across the major task categories of MMSI-Video-Bench. “SC.”, “MU.”, “Plan.” and “Pred.” refer to Spatial Construction, Motion Understanding, Planning, and Prediction, respectively.}
\label{tab:more_models}
\vspace{-1em}
\end{table*}

As shown in Tab.\ref{tab:more_models}, compared with their respective base models, only SpaceQwen exhibits almost no change in performance, whereas both Spatial-MLLM and VLM3R suffer from degradation, particularly in instruction-following ability, leading to an overall drop in performance. This trend is consistent with observations reported in prior work~\cite{yang2025mmsi,lin2025ostbench}: although such models may perform well on specific spatial reasoning datasets, their capabilities do not generalize effectively to other benchmarks and may even impair original abilities. These findings further highlight the challenge posed by our benchmark and its emphasis on comprehensively assessing models’ spatial intelligence.
\begin{table}[t]
\centering
\footnotesize
\
\begin{tabular}{l|ccc}
\toprule
Method & GPT-4o & QwenVL2.5-72B & InternVL3-78B \\
\midrule
Uniform-50 & 31.6 & 32.7 & 32.5 \\
AKS-50     & 28.4 {\scriptsize(\textcolor{red!100!black}{-3.2})} 
           & 31.9 {\scriptsize(\textcolor{red!100!black}{-0.8})} 
           & 31.6 {\scriptsize(\textcolor{red!100!black}{-0.9})} \\
\bottomrule
\end{tabular}
\vspace{-0.8em}
\caption{Comparison of model performance under uniform sampling and AKS sampling strategies.}
\label{tab:AKS_result}
\end{table}

\section{Frame Sampling Study}
\begin{figure}
  \centering
  \includegraphics[width=1.0\linewidth]{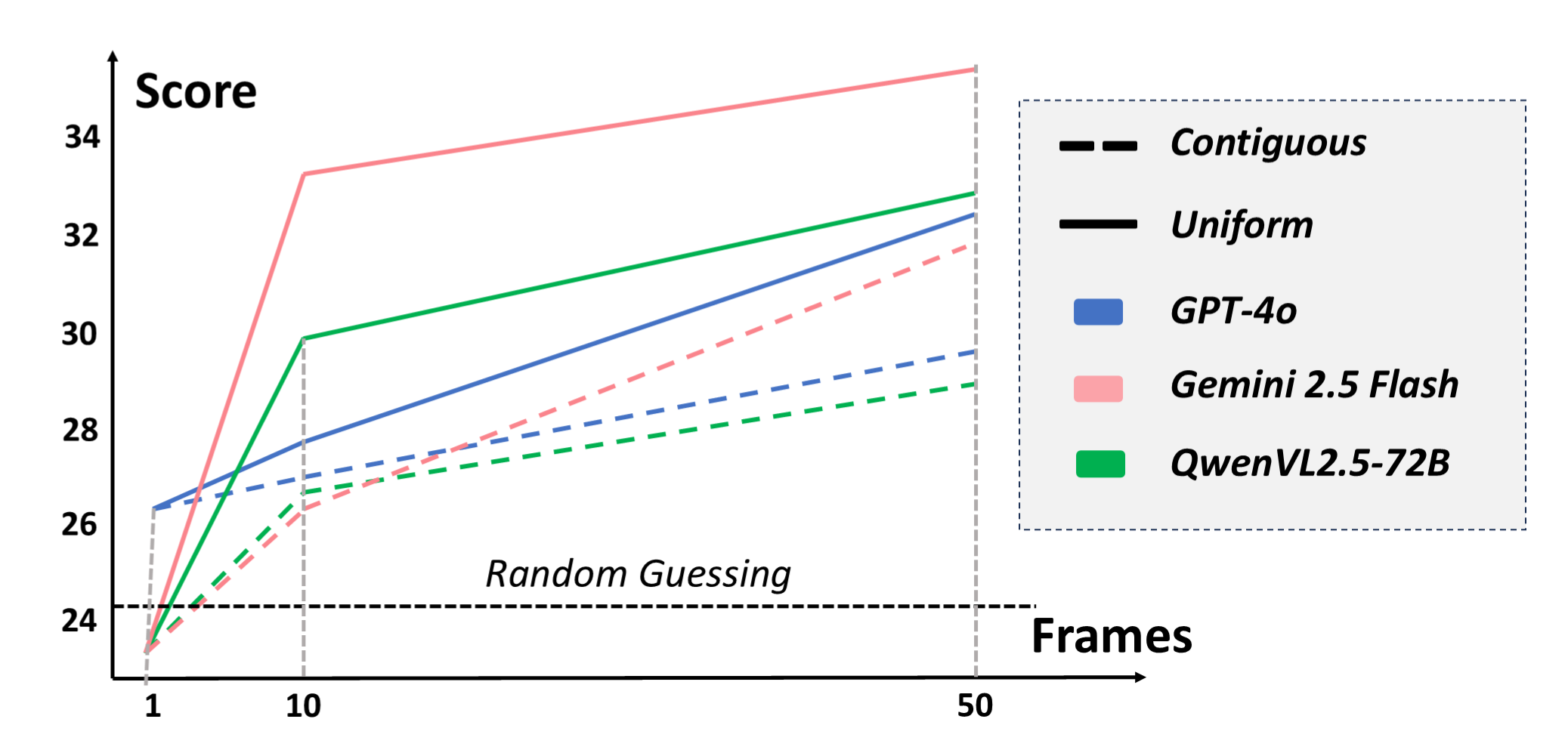}
  \vspace{-3ex}
  \caption{Model performance under different frame sampling methods. Dashed lines indicate contiguous sampling, while solid lines indicate uniform sampling.}
  \vspace{-3ex}
  \label{fig:frame analysis}
\end{figure}

\textbf{Effect of Frame Count and Sampling Strategy.} We evaluated model performance with varying frame counts (1, 10, and 50) and two sampling strategies: consecutive frames from a local segment versus uniformly sampled frames across the entire video. Experiments were conducted on three representative models (GPT-4o, Gemini 2.5 Flash, and QwenVL2.5-72B), with results shown as six curves in Fig.\ref{fig:frame analysis}.
The results reveal two key findings: (1) performance is very low at minimal frame counts, sometimes near random guessing level, indicating that there are no shortcuts in {\bench}; performance improves significantly as more frames are sampled, showing the necessity of visual information in {\bench}. (2) Uniform sampling substantially outperforms consecutive sampling, demonstrating that broad temporal coverage is essential to capture key events, and that short continuous segments are insufficient. This underscores that {\bench} is designed to require models to integrate information across the full temporal span of the video.

\noindent\textbf{Smarter Keyframe Sampling Strategy.} Recent studies have proposed more efficient frame sampling strategies that can yield notable performance improvements. Following the Adaptive Keyframe Sampling (AKS) approach~\cite{tang2025adaptivekeyframesamplinglong}, which selects frames based on image–text semantic representations, we sampled 50 frames per video and evaluated model performance. Results are summarized in Table \ref{tab:AKS_result}.
Although AKS achieves substantial gains on benchmarks such as LongVideoBench~\cite{longvideobench} and Video-MME~\cite{videomme}, it fails to provide improvements on MMSI-Video-Bench. This may be because the key frames required to answer questions in MMSI-Video-Bench cannot be directly determined from semantic similarity alone (e.g., “How does the dog move during the period when it is out of my sight?”). Relying solely on semantic cues may even narrow the model’s effective field of view, causing it to miss other critical frames.  This outcome underscores the challenging nature of our bench and indicates that it places stricter requirements on frame sampling strategies than existing video benchmarks.

\section{Error Analysis}

\subsection{Error Categorization} 

Effective video understanding requires a sequence of reasoning steps: first perceiving fine-grained details, then linking entities across frames, followed by modeling spatial relations, and finally correctly aligning prompts to answer questions, with some cases demanding deeper reasoning over implicit cues. Based on this structured process, we categorize all model errors in our bench into non-overlapping, comprehensive types (Fig.\ref{fig:error case}):
\begin{figure*}
  \centering
  \includegraphics[width=1.0\linewidth]{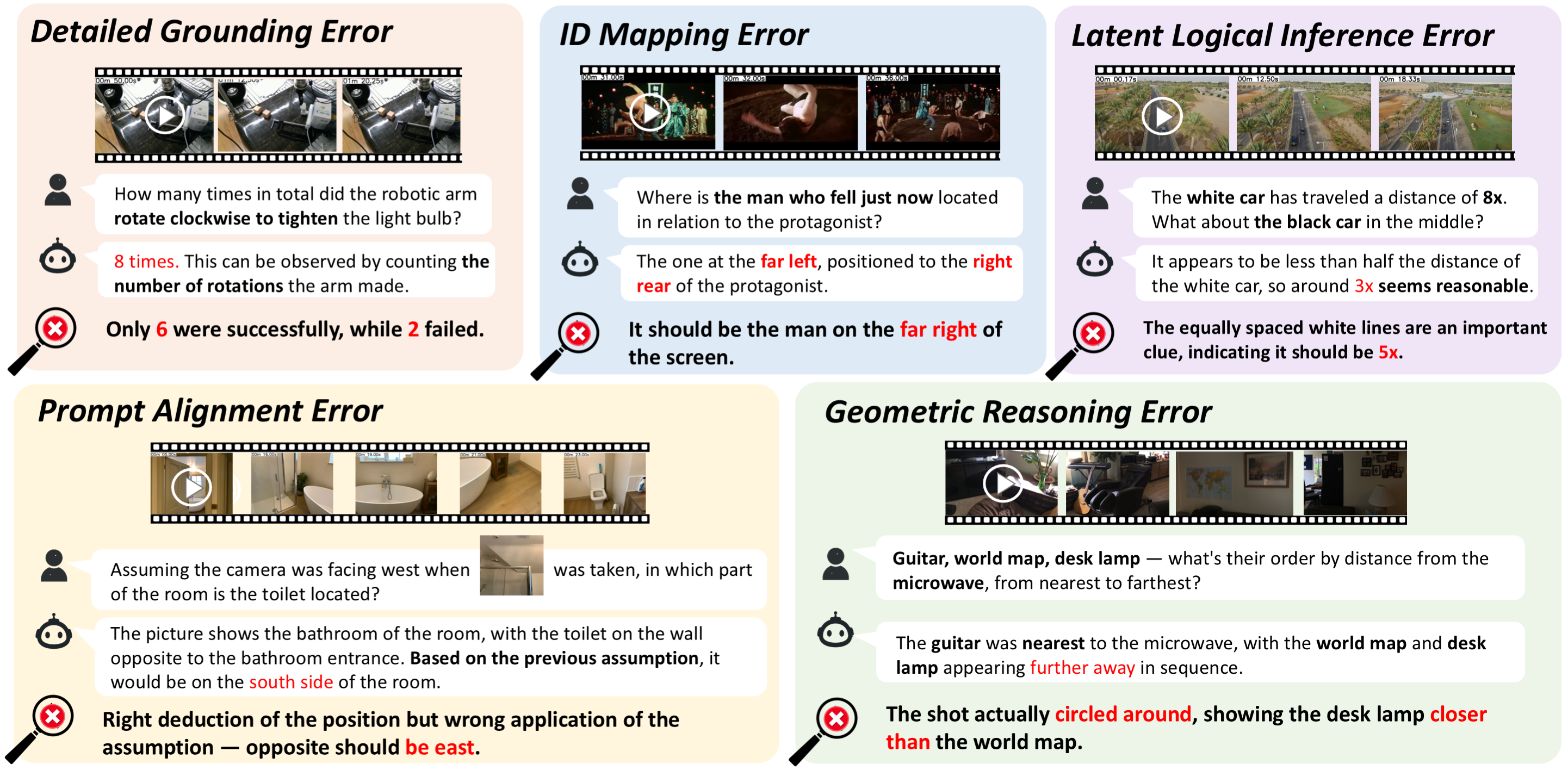}
  \vspace{-2em}
  \caption{Illustration of five representative error types identified in {\bench}, along with examples of model responses and corresponding error analyses.}
  \vspace{-3ex}
  \label{fig:error case}
\end{figure*}

\noindent \textbf{Detailed Grounding Error.} Failures in fine-grained perception, including missing or confusing objects, overlooking subtle temporal changes, or misidentifying events at specific timestamps. This error mainly reflects deficiencies in surface-level visual grounding.

\noindent \textbf{ID Mapping Error.} Failures in maintaining consistent identity tracking across frames, often caused by occlusion, rapid motion, or visually similar distractors, leading the model to confuse or mismatch entities over time.

\noindent \textbf{Geometric Reasoning Error.} Mistakes in inferring spatial relations (relative positions or distance, e.g., front/behind, near/far), revealing the model’s inability to establish coherent spatial associations across frames.

\noindent \textbf{Prompt Alignment Error.} Misunderstandings in interpreting the prompt or integrating it with visual evidence. These occur when the prompt introduces new conditions, reference images, or auxiliary visual inputs that the model fails to correctly incorporate, even if its understanding of the video information itself is accurate.

\noindent \textbf{Latent Logical Inference Error.}  Failures in reasoning that require integrating implicit cues or commonsense knowledge. Some questions in {\bench} demand inference based on subtle contextual clues, such as choosing an appropriate reference object to estimate height/ distance/ speed or correlating information across different viewpoints, or predicting motion trajectories using basic physical intuition. The model fails to detect or leverage these implicit cues.

\subsection{Error Statistics} 

We select four representative models (GPT-4o, Gemini 2.5 Flash, O3, and QwenVL2.5-72B) and conducted an error analysis on a total of 520 incorrectly answered cases, evenly sampled across different question categories. The errors were categorized and quantified, and the final statistics, shown in Fig. \ref{fig:error distribution}, illustrate the distribution of each error type within the main categories, as well as the overall composition of error types. Several observations can be made:

\begin{figure}
  \centering
  \includegraphics[width=1.1\linewidth]{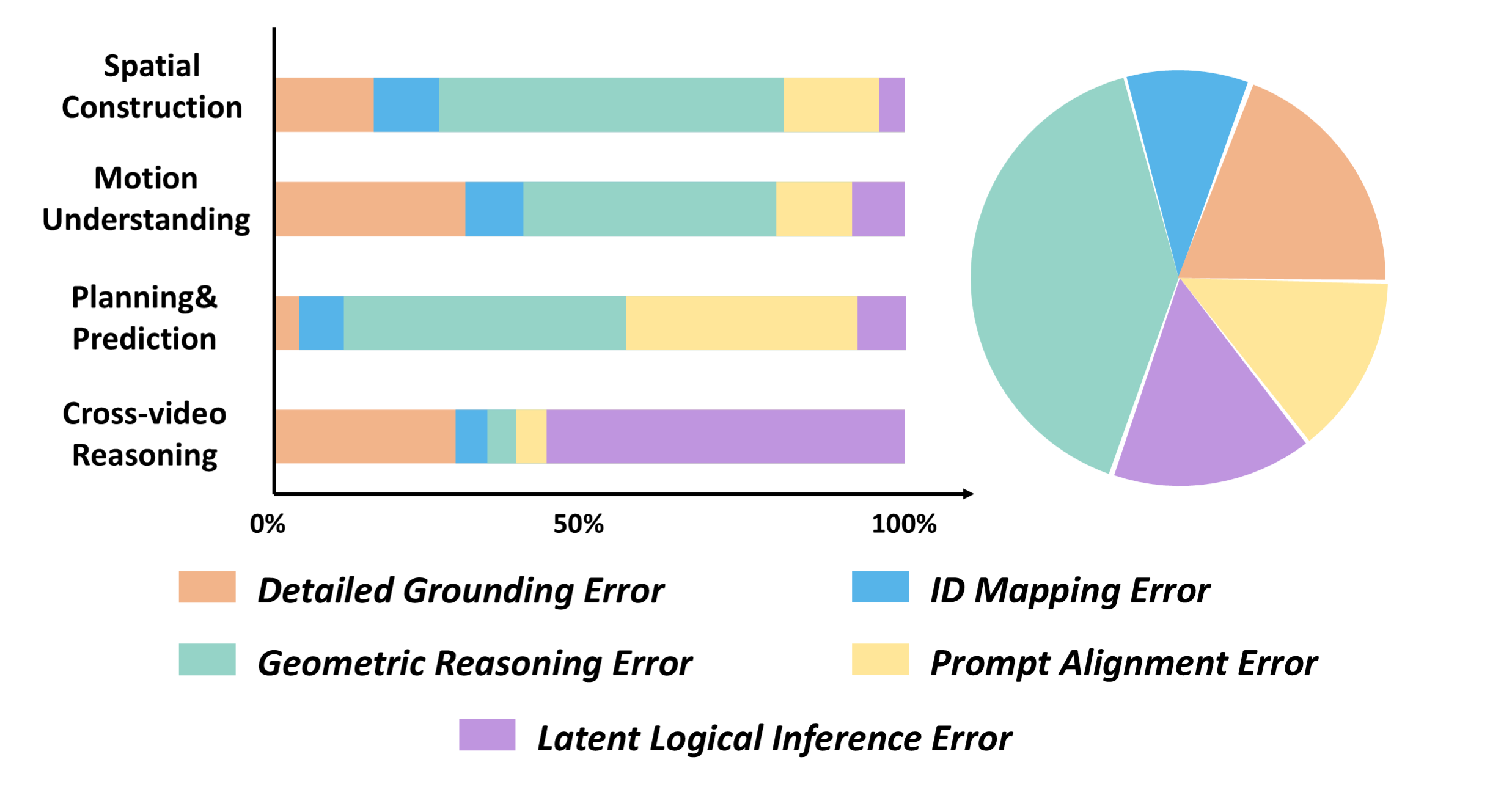}
  \vspace{-2.6em}
  \caption{Distribution of the five error types. (Left) Error distribution across different question categories. (Right) Overall proportion of each error type.}
  \vspace{-3ex}
  \label{fig:error distribution}
\end{figure}

Geometric Reasoning Error is the most prevalent error type overall, especially within the \textbf{Spatial Construction tasks}. This finding is consistent with prior spatial reasoning benchmarks\cite{yang2024think,lin2025ostbench,yang2025mmsi}, indicating that current models still struggle with inferring even simple geometric relations.

 Distinct error distributions reveal task-specific capability bottlenecks. Beyond Spatial Construction, we observe the following patterns across other task categories:

\begin{itemize}
    \item \textbf{In Motion Understanding tasks}, detailed grounding remains a major limitation: models often fail to comprehensively detect or interpret motion patterns, especially when confronted with fast movements, subtle actions, or long-duration motions.
    \item \textbf{In Planning and Prediction tasks}, Prompt Alignment Error is a significant source of issues: models may accurately perceive the spatiotemporal context but still fail to connect high-level goals, assumptions, or contextual conditions with the video evidence.
    \item  \textbf{In Cross-Video Reasoning tasks}, Latent Logical Inference Errors are most prominent, followed by Detailed Grounding Errors. These tasks typically require identifying correspondences across multiple videos (i.e., using matching instances across videos from different time points or viewpoints to establish spatio-temporal correspondences between the videos). We find that models frequently either fail to locate the same instance in both videos simultaneously or neglect to utilize them effectively for reasoning.

\end{itemize}

Through this error analysis, we gain a deeper understanding of the specific failure modes associated with each category in MMSI-Video-Bench, offering valuable insights into which model capabilities require improvement and which weaknesses future iterations should target.

  \begin{figure}
  \centering
  \includegraphics[width=1.0\linewidth]{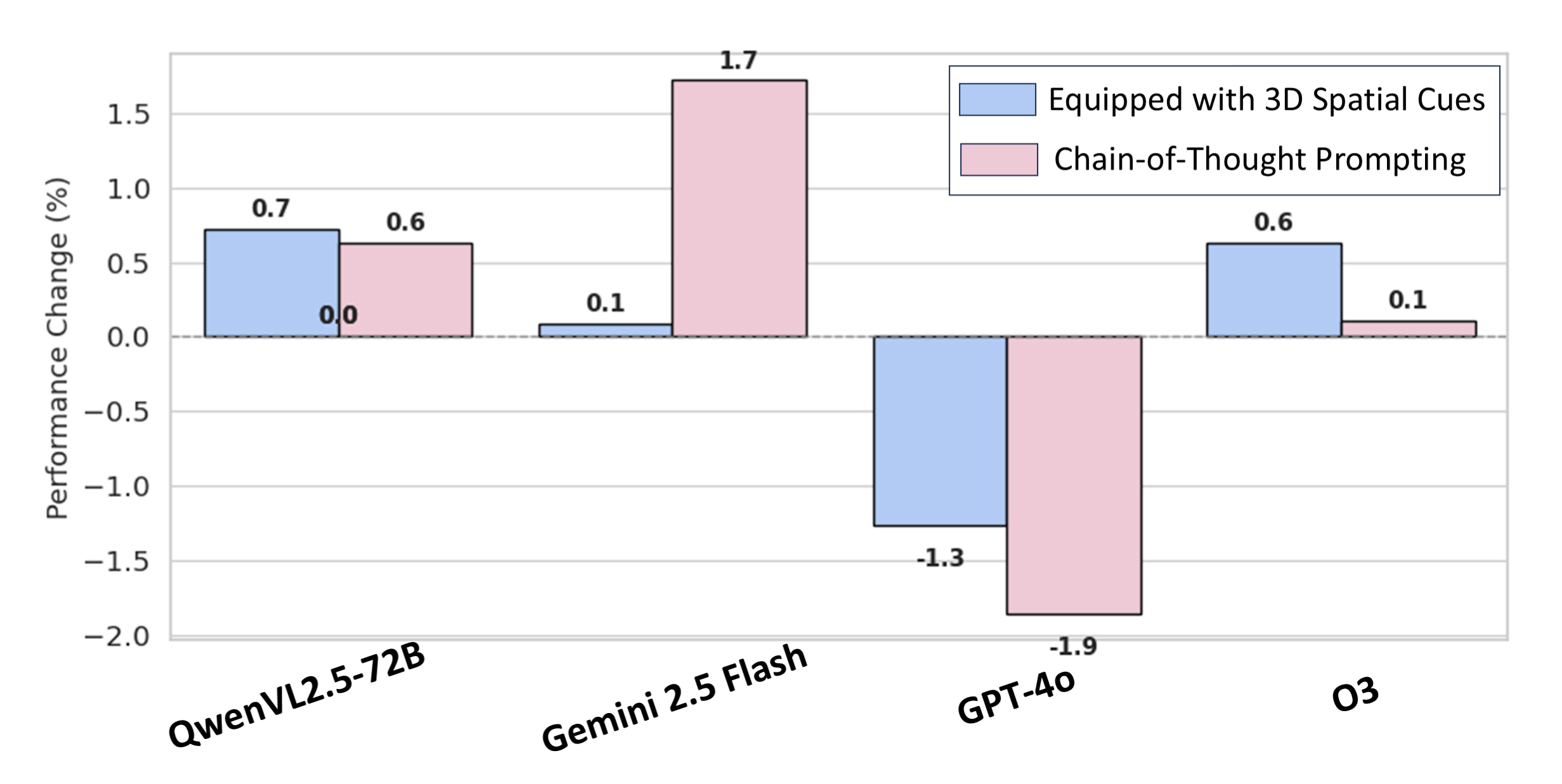}
  \vspace{-3ex}
  \caption{Effect of different methods on model performance.}
  \vspace{-3ex}
  \label{fig:try_results}
\end{figure}

\begin{figure*}
  \centering
  \includegraphics[width=1.0\linewidth]{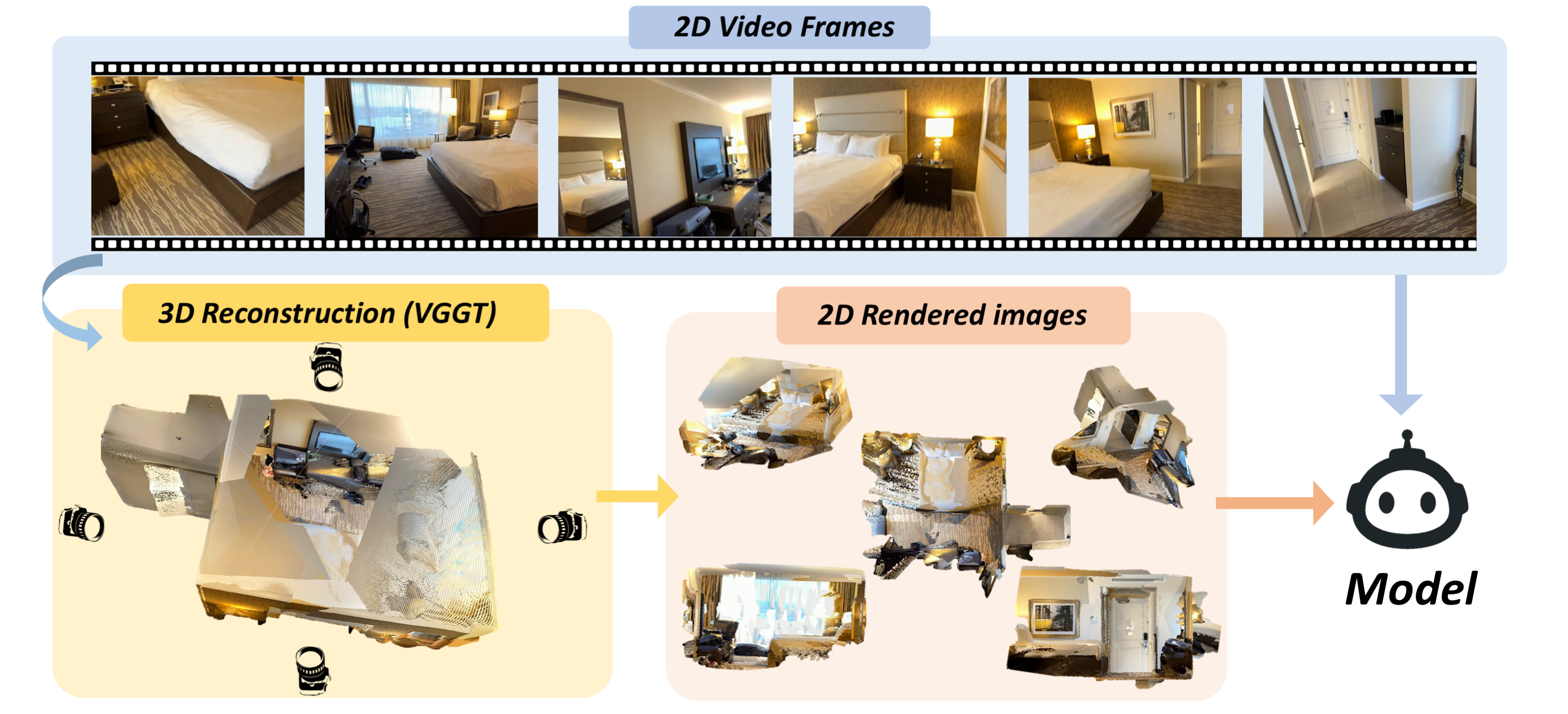}
  \vspace{-3ex}
  \caption{Pipeline of equipping the model with 3D spatial cues.}
  \label{fig:try}
\end{figure*}

\vspace{3ex}
\section{Preliminary Exploration for Model Improvement}

While our error analysis categorizes and quantifies the types of failures made by existing models, in this section, we conduct an initial exploration toward improving model performance based on these identified errors.
  
\subsection{Equipping Models with 3D Spatial Cues} 

Among all error types, Geometric Reasoning Error stems from the model’s insufficient ability to build and utilize spatial representations. Correspondingly, we consider two general directions for improving spatial reasoning: (1) training models with sufficient and diverse spatial reasoning data, and (2) enhancing models by explicitly providing or modeling spatial representations, e.g., through dedicated architectures or auxiliary tools.
  In our preliminary attempt, we adopt the second strategy. Specifically, we equip the model with spatial cues generated by VGGT~\cite{wang2025vggtvisualgeometrygrounded}, enabling it to better perceive global scene geometry. As illustrated in Fig.\ref{fig:try}, we first feed raw video frames into VGGT to obtain a 3D reconstruction of the scene. We then render 10 multi-view observations (including top-down and multiple side views) from the reconstructed point cloud. These sparse geometric cues are combined with the original video frames and fed into the model together as input.

We evaluate four representative models: Gemini 2.5 Flash, O3, GPT-4o, and QwenVL2.5-72B. Under the Uniform-50 setting, 50 frames from each video were fed into VGGT for 3D reconstruction and rendered into corresponding images. After equipping the models with 3D spatial cues, their performance is shown in Fig.\ref{fig:try_results}. All four models show no significant improvement (with gains below 1\%), suggesting that 3D spatial cues do not reliably enhance spatial intelligence under our current setup. Upon further analysis of model errors, we identified two main issues:
\begin{itemize}
    \item Issues in generating 3D spatial cues. While VGGT can handle relatively simple scenes, such as indoor scanning, it often fails in complex scenarios involving multi-room or multi-floor scans or dynamic scenes. In these failure cases, the rendered images provide little to no useful information for the models and may even introduce noise. This reflects an inherent limitation of VGGT; to consistently provide accurate 3D spatial cues, more robust and generalizable tools are needed.
    \item Issues in utilizing 3D spatial cues. Examination of the models’ reasoning processes revealed that the models fail to effectively leverage the 3D spatial cues. In many cases, the cues are either ignored or not correctly associated with the video content and the question, even though our prompts explicitly instructed the model to use them. This indicates that designing spatial cues that are easily interpretable by the models remains an open challenge.
\end{itemize}

\subsection{Chain-of-Thought Prompting.} 

To address issues such as Prompt Alignment and Latent Logic Inference errors, we explored Chain-of-Thought (CoT) prompting, guiding models to reason step by step. The model is provided with explicit prompts for each step: 
\begin{itemize}
    \item \textit{Step1: Understand and Analyze. Interpret the problem input, including auxiliary visual inputs, preset conditions, or requirements, and identify the key information to extract from the video.}
    \item \textit{Step2: Locate and Gather Evidence. Find the relevant information in the video and collect sufficient evidence, including implicit clues not directly mentioned in the input.}
    \item \textit{Step3: Reason and Solve. Combine the prompt with the extracted video information and perform step-by-step reasoning to answer the question.}
\end{itemize}

As shown in Fig.\ref{fig:try_results}, simply encouraging the model to “think step by step” does not consistently improve performance. This aligns with previous findings~\cite{yang2024think,yang2025mmsi}. The underlying issue is not that the model forgets to perform certain steps, but rather that it struggles to handle inherently difficult aspects of the task, highlighting that the limitation lies in the model’s intrinsic reasoning ability.

\section{Additional Perspectives of MMSI-Video-Bench}
Due to the diversity of data sources and the holistic coverage of task types in MMSI-Video-Bench, the benchmark can be also examined from several domain-oriented perspectives. Based on different application focuses, we further derive three subset benchmarks from MMSI-Video-Bench and report model performance on each of them. Similarly, model performance is evaluated under both the Uniform-50 and Sufficient-Coverage settings.

\begin{table*}[tbp!]
  \centering
  \scriptsize
  \setlength{\tabcolsep}{1.3pt}
  \begin{adjustbox}{width=\textwidth,center}
    \begin{tabularx}{\textwidth}{%
      l
      *{4}{>{\centering\arraybackslash}p{0.100\textwidth}}  
      *{3}{>{\centering\arraybackslash}p{0.070\textwidth}}  
      *{3}{>{\centering\arraybackslash}p{0.070\textwidth}}  
    }
    \toprule
    \multirow{2}{*}{\textbf{Model}}
      & \multicolumn{4}{c}{\cellcolor{cyan!8}\textbf{Indoor Scene Perception}}
      & \multicolumn{3}{c}{\cellcolor{yellow!30}\textbf{Robot}}
      & \multicolumn{3}{c}{\cellcolor{green!15}\textbf{Grounding}}\\
    & \scriptsize Avg. & \scriptsize Static-CC. & \scriptsize Static-IC.
    & \scriptsize Dynamic & \scriptsize Avg. & \scriptsize Man.
    & \scriptsize Nav. & \scriptsize Avg. & \scriptsize TG.
    & \scriptsize TL. \\
    \midrule
    \rowcolor{gray!10}\multicolumn{11}{l}{\emph{\textbf{(Sufficient-Coverage)  Proprietary}}} \\

O4-mini &37.1 &38.2 &33.5 &\textbf{49.2} &35.3 &28.9 &\textbf{41.1} &31.3 &32.4 &25.9\\
O3 &\textbf{39.4} &37.6 &38.2 &\textbf{49.2} &36.3 &\textbf{38.1} &34.6 &37.6 &38.1 &35.2\\
GPT-4o &29.6 &28.0 &29.0 &36.9 &28.9 &25.8 &31.8 &29.2 &30.2 &24.1\\
Gemini 2.5 Flash &\textbf{39.4} &\textbf{39.8} &\textbf{39.0} &40.0 &35.8 &\textbf{38.1} &33.6 &37.6 &\textbf{39.9} &25.9\\
-Thinking &37.9 &36.6 &37.1 &44.6 &\textbf{37.2} &35.0 &39.2 &\textbf{38.8} &38.8 &\textbf{38.9}\\
   
   \rowcolor{gray!10}\multicolumn{11}{l}{\emph{\textbf{(Sufficient-Coverage)  Open-source}}} \\
   
InternVL2.5-8B &28.7 &26.3 &27.9 &\textbf{38.5} &27.9 &33.0 &23.4 &30.1 &29.5 &33.3\\
InternVL3-8B &27.7 &24.2 &30.5 &26.1 &29.9 &33.0 &27.1 &30.8 &\textbf{30.6} &31.5\\
InternVideo2.5-8B &26.8 &24.2 &26.5 &35.4 &27.9 &25.8 &29.9 &27.8 &28.5 &24.1\\
QwenVL2.5-7B &24.5 &24.2 &24.6 &24.6 &26.5 &29.9 &23.4 &28.7 &28.1 &31.5\\
QwenVL2.5-32B &29.6 &\textbf{30.1} &30.5 &24.6 &32.8 &32.0 &33.6 &\textbf{31.0} &29.9 &\textbf{37.0}\\
QwenVL2.5-72B &29.2 &26.3 &29.0 &\textbf{38.5} &\textbf{37.8} &40.2 &\textbf{35.5} &30.4 &\textbf{30.6} &29.6\\
QwenVL3-8B &27.9 &24.3 &29.6 &31.1 &32.1 &\textbf{40.2} &24.8 &26.5 &26.8 &25.0\\
QwenVL3-30B &\textbf{30.0} &29.0 &29.0 &36.9 &32.8 &33.0 &32.7 &28.1 &28.1 &27.8\\
-Thinking &29.8 &26.9 &\textbf{32.7} &26.1 &27.9 &32.0 &24.3 &26.9 &25.3 &35.2\\

\rowcolor{gray!10}\multicolumn{11}{l}{\emph{\textbf{(Uniform-50)  Proprietary}}} \\
Claude-haiku-4.5 &33.5 &34.4 &32.4 &35.4 &34.8 &39.2 &30.8 &32.8 &32.0 &37.0\\
GPT-5 &\textbf{41.7} &40.3 &\textbf{42.6} &41.5 &37.8 &39.2 &36.5 &35.2 &35.6 &33.3\\
O4-mini &37.5 &40.3 &36.0 &35.4 &33.3 &37.1 &29.9 &34.3 &33.5 &\textbf{38.9}\\
O3 &40.7 &43.0 &38.2 &\textbf{44.6} &39.2 &36.1 &\textbf{42.1} &37.3 &37.7 &35.2\\
GPT-4o &31.7 &38.2 &26.5 &35.4 &29.9 &28.9 &30.8 &31.9 &31.3 &35.2\\
Gemini 3 Pro &39.4 &36.0 &40.8 &43.1 &\textbf{40.2} &38.1 &\textbf{42.1} &35.2 &35.6 &33.3\\
Gemini 2.5 Flash &39.2 &\textbf{44.6} &36.8 &33.9 &33.8 &38.1 &29.9 &\textbf{38.2} &\textbf{38.1} &\textbf{38.9}\\
-Thinking &36.7 &39.8 &33.8 &40.0 &39.7 &39.2 &40.2 &36.1 &36.3 &35.2\\
Doubao-1.5-thinking &33.0 &30.8 &32.5 &41.9 &36.1 &\textbf{41.4} &31.2 &37.0 &37.1 &36.8\\
Seed-1.6-vision &34.2 &36.1 &31.9 &37.2 &39.3 &\textbf{41.4} &37.5 &33.0 &33.3 &31.6\\
\rowcolor{gray!10}\multicolumn{11}{l}{\emph{\textbf{(Uniform-50)  Open-source}}} \\

InternVL2.5-8B &29.4 &25.8 &30.1 &36.9 &28.4 &28.9 &28.0 &28.4 &28.8 &25.9\\
InternVL2.5-38B &28.3 &24.7 &27.9 &\textbf{40.0} &36.3 &36.1 &\textbf{36.5} &31.9 &31.3 &35.2\\
InternVL2.5-78B &30.4 &28.5 &31.6 &30.8 &34.8 &33.0 &\textbf{36.5} &29.9 &29.2 &33.3\\
InternVL3-8B &27.0 &22.0 &29.4 &30.8 &\textbf{37.8} &39.2 &\textbf{36.5} &31.9 &31.0 &\textbf{37.0}\\
InternVL3-38B &29.1 &24.7 &31.6 &30.8 &27.9 &28.9 &27.1 &30.4 &31.0 &27.8\\
InternVL3-78B &\textbf{32.5} &28.5 &\textbf{34.6} &35.4 &34.3 &38.1 &30.8 &\textbf{35.5} &\textbf{35.2} &\textbf{37.0}\\
InternVideo2.5-8B &26.8 &26.9 &26.5 &27.7 &29.9 &28.9 &30.8 &27.2 &25.6 &35.2\\
LLaVA-Video-7B &27.5 &22.6 &28.3 &38.5 &24.5 &29.9 &19.6 &27.2 &27.1 &27.8\\
LLaVA-Video-72B &28.1 &20.4 &32.0 &33.9 &34.3 &39.2 &29.9 &31.0 &31.3 &29.6\\
QwenVL2.5-7B &27.1 &26.9 &28.7 &21.5 &34.8 &35.0 &34.6 &26.6 &25.6 &31.5\\
QwenVL2.5-32B &26.6 &24.2 &26.1 &35.4 &30.4 &33.0 &28.0 &27.5 &26.0 &35.2\\
QwenVL2.5-72B &30.8 &29.0 &29.8 &\textbf{40.0} &34.8 &\textbf{46.4} &24.3 &34.3 &34.2 &35.2\\
QwenVL3-8B &28.7 &27.4 &27.6 &36.9 &27.0 &30.9 &23.4 &28.7 &28.1 &31.5\\
QwenVL3-30B &27.5 &23.1 &29.4 &32.3 &32.8 &35.0 &30.8 &29.2 &28.1 &35.2\\
-Thinking &32.3 &\textbf{29.6} &33.5 &35.4 &27.9 &29.9 &26.2 &31.6 &30.6 &\textbf{37.0}\\

    \bottomrule
    \end{tabularx}
  \end{adjustbox}
  \vspace{-1.1em}
  \caption{Performance of different models on the three subset benchmarks, including overall results and scores for each task subtype. “IC” and “CC” denote Instance-Centric and Camera-Centric, respectively; “Man.” and “Nav.” represent Manipulation and Navigation; and “TG” and “TL” refer to Target Grounding and Time Localization, respectively.}
  \label{tab:other_bench}
  \vspace{-1em}
\end{table*}

\begin{figure*}
  \centering
  \includegraphics[width=0.9\linewidth]{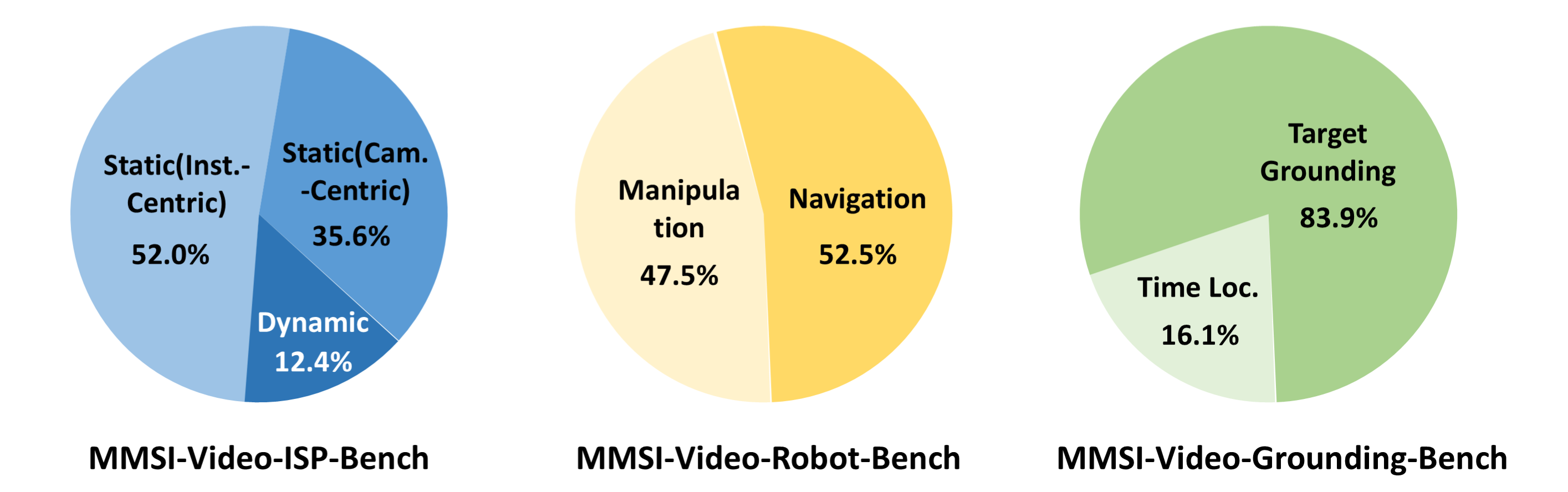}
  \vspace{-1ex}
  \caption{Distribution of subtask proportions across the Indoor Scene Perception Bench, Robot Bench and Grounding Bench.}
  \label{fig:other dis}
\end{figure*}

\noindent\textbf{Indoor Scene Perception Bench.} The Indoor Scene Perception Bench focuses on evaluating a model’s ability to perceive and understand indoor environments. This subset contains 523 samples from MMSI-Video-Bench and includes three major categories: Static-Scene (Instance-Centric), Static-Scene (Camera-Centric), and Dynamic-Scene. The two static-scene categories assess a model’s understanding of static indoor layouts. The Instance-Centric category includes questions that are independent of the camera or viewer perspective, targeting object-intrinsic spatial attributes and inter-object spatial relations within the scene. In contrast, the Camera-Centric category examines spatial relations defined relative to the viewer or camera, evaluating the model’s understanding of its positional relationship to the surrounding environment. The Dynamic-Scene category tests a model’s ability to reason about scene changes over time, including those caused by human activities as well as object replacement events that occur between temporally separated video segments.

  The evaluation results, shown in the left four columns of Tab.\ref{tab:other_bench}, indicate that among all models, GPT-5 achieves the strongest performance; notably, GPT-5 excels in instance-centric static scene perception, while Gemini 2.5 Flash achieves the best performance in camera-centric static scene perception. In addition, O3 and O4-mini show a strong capability in understanding scene changes. Most open-source models lag behind the proprietary ones. Looking across sub-tasks, models with strong overall performance tend to maintain balanced scores across all types, whereas weaker models exhibit their primary bottleneck in Static-CC, which requires reasoning about the spatial relationship between the observer and the environment—a capability that these models struggle with.

\noindent\textbf{Robot Bench.} The Robot Bench focuses on evaluating model performance on two core tasks in real-world embodied scenarios: Manipulation and Navigation. This subset contains 204 samples. The Manipulation category assesses a model’s ability to perceive and reason about fine-grained tabletop operations and interactive motions, while the Navigation category evaluates a model’s planning and navigation capabilities within indoor environments.

As reported in the middle three columns of Tab.\ref{tab:other_bench}, Gemini 3 Pro stands out with the strongest overall results on this benchmark. Performance on individual subtasks: QwenVL2.5-72B delivers the best results on Manipulation, while O3 and Gemini 3 Pro lead on Navigation. Notably, the Navigation task reveals substantially larger performance gaps between models and highlights a key weakness of many open-source models. 

\noindent\textbf{Grounding Bench.}   The Grounding Bench comprises 335 samples and requires models to localize either target objects or specific time points within a video. Unlike traditional visual grounding or temporal localization benchmarks that mainly involve semantic referential grounding, our benchmark distinguishes itself by requiring spatial reasoning for all queries to correctly identify the target object or temporal segment. Naturally, this subset is divided into two components based on the type of grounding: target grounding and temporal localization.   Within the Grounding Bench, Gemini 2.5 Flash achieves the strongest overall performance, excelling in both temporal localization and target-object identification. O4-mini demonstrates similarly strong temporal localization capabilities as well.

These three benchmarks evaluate model performance within more fine-grained domains and categories, enabling targeted assessment of specific model capabilities. They also provide a convenient evaluation protocol for models designed for particular domains or task types.
\section{Conclusion}
We present {\bench}, a diverse, human-annotated, holistic video-based spatial intelligence benchmark that evaluates models’ perception, understanding, reasoning, and decision-making over spatiotemporal information, complemented by three domain-oriented sub-benchmarks that offer targeted perspectives.
Our evaluation reveals a substantial gap between model and human performance, with models struggling across all task categories beyond spatial construction, and even spatially fine-tuned models failing to generalize effectively to our benchmark.
Error analyses expose task-specific failure patterns that highlight concrete weaknesses in current models; our preliminary explorations further show that neither 3D spatial cues nor chain-of-thought prompting yields meaningful gains; and the frame-sampling study underscores the benchmark’s difficulty and the need for more effective sampling strategies.
Overall, {\bench} provides a rigorous and holistic testbed for assessing spatial intelligence in video models, while our analyses offer actionable insights and directions for future improvements.
\section{Acknowledgement}\label{sec:Acknowledgement} This work is funded in part by the National Key R\&D Program of China, and Shanghai Artificial Intelligence Laboratory.
\appendix
\vspace{2em}
\noindent{\large \textbf{Appendix}}

\setcounter{page}{1}
\startcontents
{
    \hypersetup{linkcolor=gray}
    \printcontents{}{1}{}
}
\section{Benchmark Details}

\subsection{Task Formulation Details}
\begin{figure*}
  \centering
  \includegraphics[width=1.0\linewidth]{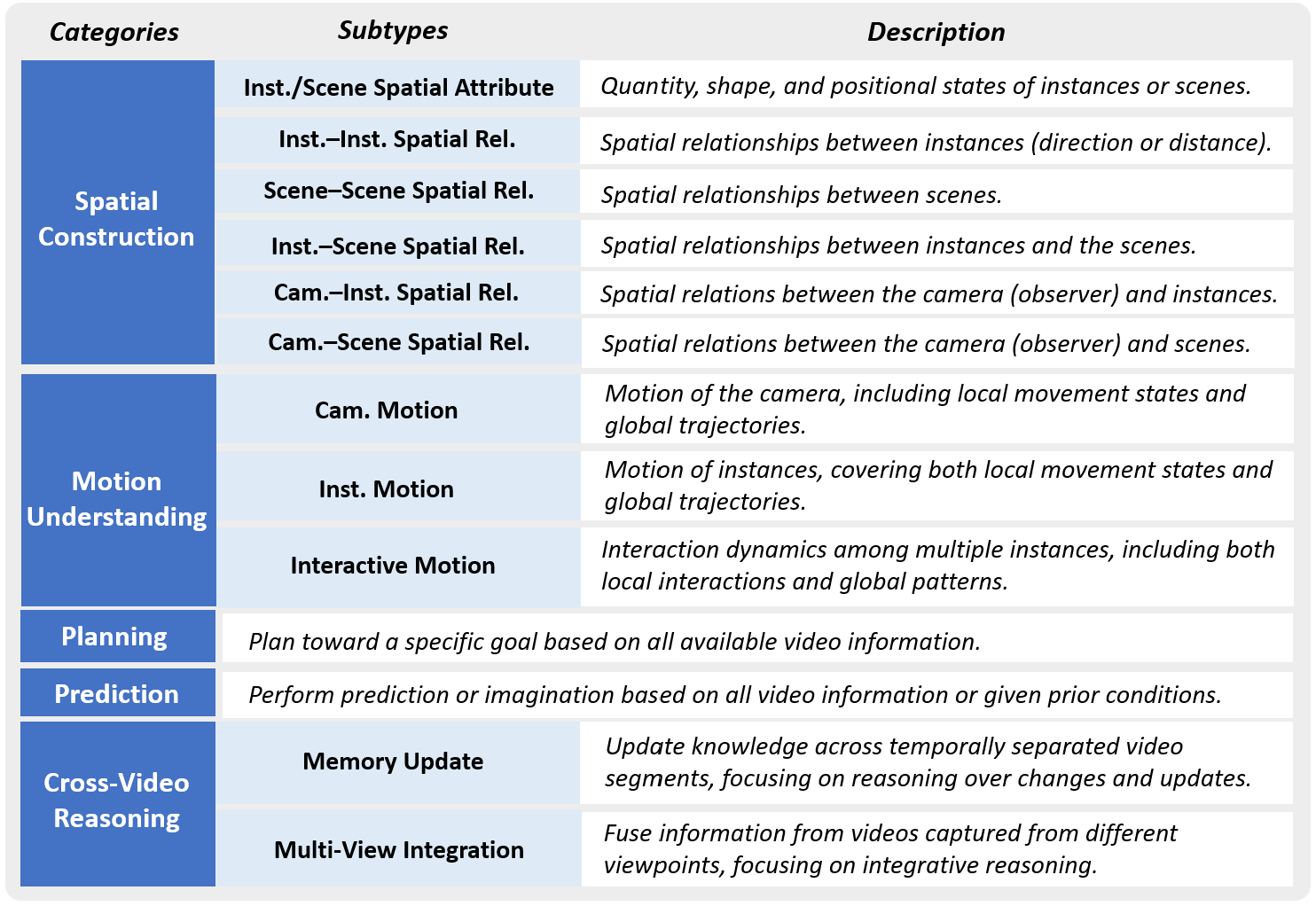}
  \caption{Task Categories and Subtypes in {\bench}. "Inst." denotes "instance", "Cam." denotes "camera" and "Rel." denotes "relationship".}
  \vspace{-3ex}
  \label{fig:type_detail}
\end{figure*}

  As defined in the main paper, MMSI-Video-Bench is organized into five main categories: Spatial Construction, Motion Understanding, Planning, Prediction, and Cross-Video Reasoning.
 Spatial Construction evaluates the spatial attributes of instances and scenes, as well as the pairwise spatial relations among instances, scenes, and the camera.
 Motion Understanding is further divided into three aspects: camera motion, instance motion, and inter-instance interactive motions.
 Cross-Video Reasoning encompasses two subtypes: Memory Update and Multi-View Integration.
  In total, MMSI-Video-Bench consists of 5 main categories and 13 subtypes, with their detailed definitions summarized in Fig.\ref{fig:type_detail}.

  \begin{table}[t]
\centering
\footnotesize
\
\begin{tabular}{l|ccc}
\toprule
Dataset & Type & FPS & Duration(Sec.) \\
\midrule
Roomtour3d~\cite{roomtour3d} & Indoor Scan. & 1.0 & 466.86  \\
ScanNet~\cite{scannet}& Indoor Scan. & 1.0 & 39.92  \\
ScanNet++~\cite{scannet++}& Indoor Scan. & 2.0 & 136.35  \\
3RScan~\cite{3rscan} & Indoor Scan.& 1.0 & 60.94  \\
ARKitScenes~\cite{arkitscenes}& Indoor Scan. & 1.0 & 77.28  \\
RealEstate10k~\cite{realestate}& Indoor Scan. & 0.66 & 214.73  \\
DL3DV~\cite{dl3dv}& Indoor\&Outdoor & 1.0 & 39.81  \\
Waymo~\cite{Waymo}& Outdoor Env. & 5.0 & 15.92  \\
NuScenes~\cite{nuScenes}& Outdoor Env. & 4.0 & 26.79  \\
OVIS~\cite{qi2022occludedvideoinstancesegmentation} &Outdoor Env.& 5.0 & 13.74  \\
TrackingNet~\cite{müller2018trackingnetlargescaledatasetbenchmark} & Outdoor Env.& 4.0 & 21.33  \\
LaSOT~\cite{LaSOT} & Outdoor Env.& 5.89 & 32.40  \\
UAV123~\cite{uav123}& Outdoor Env. & 5.89 & 24.32  \\
Ego4D~\cite{ego4d}& Ego.-Int. & 2.0/8.33 & 262.49  \\
EPIC-KITCHENS~\cite{EPICKITCHENS}& Ego.-Int.  & 2.0/8.33 & 91.51  \\
EgoExoLearn~\cite{huang2024egoexolearn}& Ego.-Int. & 4.0 & 565.81  \\
MultiSports~\cite{Li_2021_ICCV}& Exo.-HA. & 2.0/8.33 & 20.51  \\
charades~\cite{sigurdsson2016hollywood}& Exo.-HA. & 4.0 & 27.91  \\
LEMMA~\cite{lemma}& Exo.-HA. & 12.50 & 22.84  \\
TF2023~\cite{Zhao2024FusingPA}& Exo.-HA. & 2.0 & 492.46  \\
CVMHT~\cite{han2020cvmht}& Exo.-HA. & 4.0 & 37.25  \\
AVA~\cite{han2020cvmht}& Exo.-HA. & 1.0 & 900.27  \\
DROID~\cite{khazatsky2024droid}& Others & 4.0 & 86.75  \\
RH20T~\cite{fang2024rh20t} & Others & 4.0 & 84.32  \\
DTU~\cite{jensen2014large}& Others & 2.0 & 24.00  \\
RealWorld & Indoor\&Outdoor & 2.0 & 46.43  \\
\bottomrule
\end{tabular}
\vspace{-0.8em}
\caption{Statistics of all source video datasets, including their capture types, average durations, and standardized FPS after pre-processing.Scan./Env. denotes scanning, environment.; Exo.-Int. denotes egocentric interactions and Exo.-HA. denotes exocentric human activities.}
\label{tab:video_detail}
\vspace{-2.3em}
\end{table}

\subsection{Data Collection \& Preprocessing Details}
Our benchmark is constructed from 25 publicly available video datasets, complemented by additional videos captured and collected by ourselves. During the data preprocessing, we perform filtering to remove clips that are too short in duration, and for datasets originally provided in frame format, we reconstruct videos by concatenating frames according to the FPS specified in their corresponding papers.

Each video dataset falls into one of several capture types, including indoor scanning, outdoor environment, egocentric interactions, exocentric human activities, and other categories. As mentioned in the main paper, we standardize the frame rate for each video category to an appropriate value that ensures no key information is lost. The capture types, frame-rate settings, and average duration statistics for each category are summarized in Tab.~\ref{tab:video_detail}.

\subsection{Human Annotation UI}

\begin{figure*}
  \centering
  \includegraphics[width=1.0\linewidth]{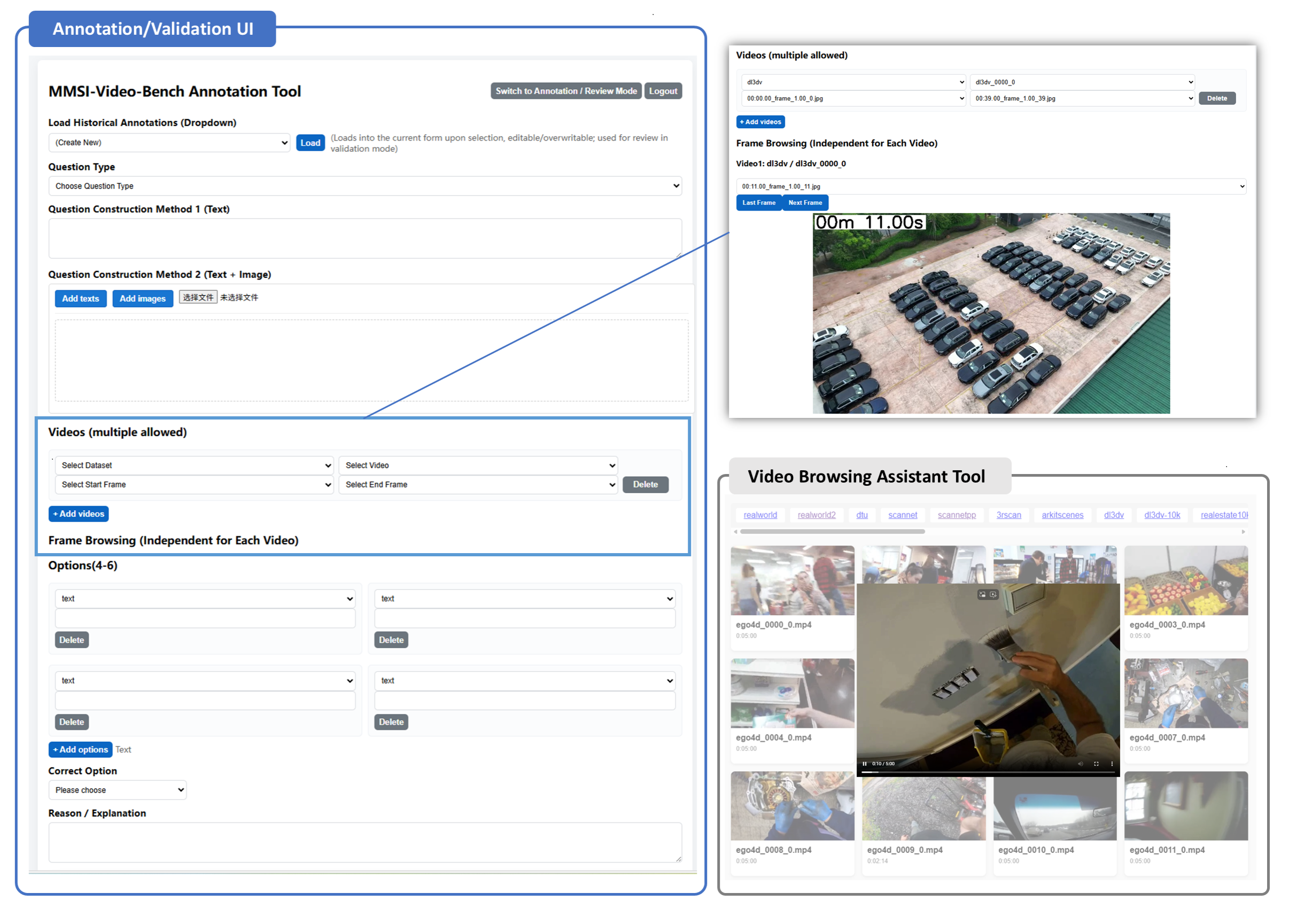}
  \vspace{-3ex}
  \caption{User interface for annotation and quality validation in MMSI-Video-Bench.}
  \vspace{-3ex}
  \label{fig:UI}
\end{figure*}

  As shown in Fig.\ref{fig:UI}, we provide a dedicated UI tool for both annotation and validation.
 The interface allows users to switch between Annotation Mode and Validation Mode. In the annotation mode, annotators can select a question type, choose the corresponding video, and determine appropriate start/end frames to construct a question. The system supports questions in either pure text or text+image format. Annotators then design the answer options, assign the correct answer, and provide the reasoning behind it. The UI clearly displays timestamps corresponding to each frame, helping annotators position temporal cues precisely. Additionally, to improve annotation efficiency, we provide a Video Browsing Assistant Tool, enabling quick coarse-level navigation and preview of video content.
 
  In the validation mode, validators are randomly assigned samples annotated by others. They can inspect the loaded annotation and choose to Accept or Reject it. For rejected samples, the validator is required to provide reasons and suggestions for revision.

\subsection{More Statistics}
\begin{figure}
  \centering
  \includegraphics[width=1.0\linewidth]{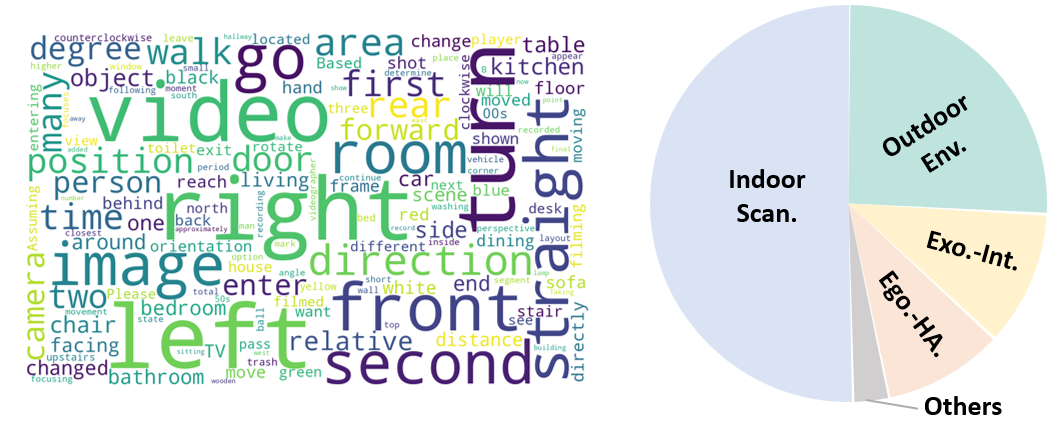}
  \vspace{-3ex}
  \caption{(Left) Word cloud of MMSI-Video-Bench. (Right) Distribution of video source categories across all samples in MMSI-Video-Bench.}
  \vspace{-3ex}
  \label{fig:dis2}
\end{figure}

   Fig.\ref{fig:dis2} illustrates a word cloud of the benchmark annotations, together with the distribution of video source capture types.

\section{Experiment Details}

\begin{figure*}
  \centering
  \includegraphics[width=1.0\linewidth]{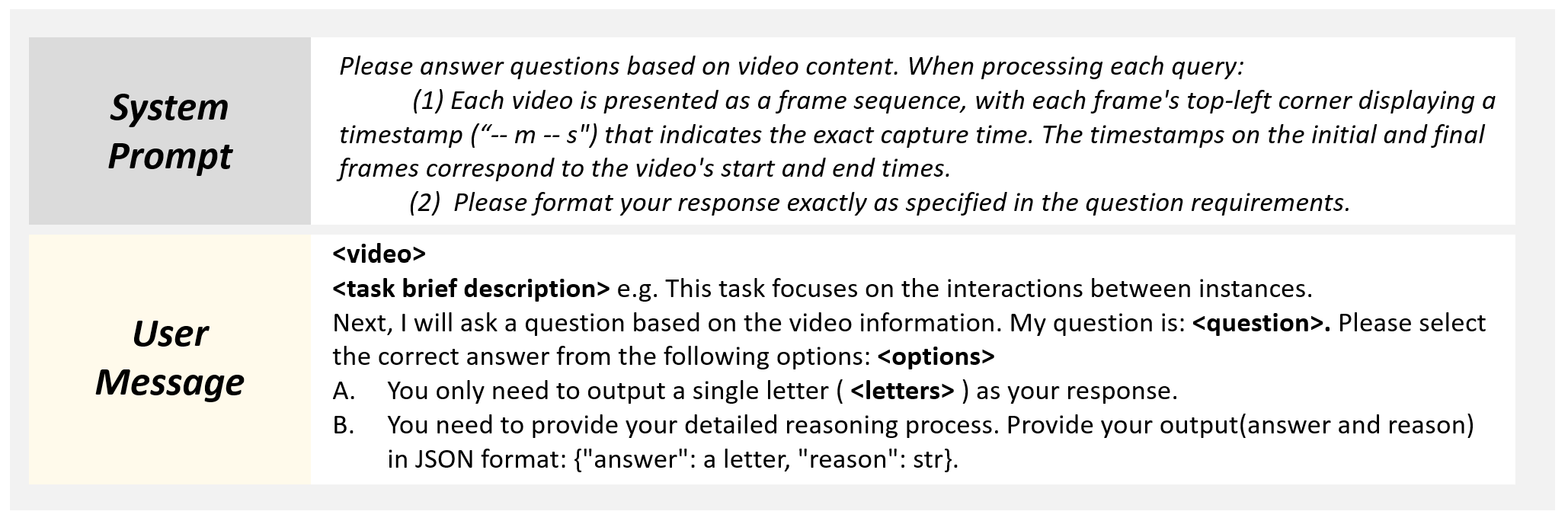}
  \caption{Structure of system and user prompts used in the experiments.}
  \vspace{-3ex}
  \label{fig:form_detail}
\end{figure*}

In our evaluation, all models are provided with the same input template. As illustrated in Fig.\ref{fig:form_detail}, the system prompt specifies the timestamp information for each frame and enforces the required output format. The user message injects, in order, the video, a brief task description, and the question with its options into the template. The expected output format adapts to the evaluation setting (e.g. During error analysis we require models to produce both an answer and a reason to facilitate failure localization, while under Chain-of-Thought prompting we require the model to emit each intermediate thinking step in addition to the final answer.)

For model outputs, we employ a general-purpose answer extraction function to parse the predicted answers. This ensures consistent extraction across different models, and we verify that all correct responses produced by any model can be accurately detected and extracted.

{
    \small
    \newpage
    \clearpage
    \bibliographystyle{ieeenat_fullname}
    \bibliography{main}
}


\end{document}